\documentclass[sigconf]{acmart}

\AtBeginDocument{%
  }

\copyrightyear{2025}
\acmYear{2025}
\setcopyright{cc}
\setcctype{by}
\setcctype{by}
\acmConference[SIGSPATIAL '25]{The 33rd ACM International Conference on
Advances in Geographic Information Systems}{November 3--6,
2025}{Minneapolis, MN, USA}
\acmBooktitle{The 33rd ACM International Conference on Advances in
Geographic Information Systems (SIGSPATIAL '25), November 3--6, 2025,
Minneapolis, MN, USA}
\acmDOI{10.1145/3748636.3764162}
\acmISBN{979-8-4007-2086-4/2025/11}

\usepackage{graphicx} 
\usepackage{multirow}
\usepackage[linesnumbered,ruled,vlined]{algorithm2e}

\title{DeepTopoNet: A Framework for Subglacial Topography Estimation on the Greenland Ice Sheets}

\author{Bayu Adhi Tama}
\affiliation{
\institution{iHARP, University of Maryland, Baltimore County (UMBC)}
\country{USA}
}
\email{bayu@umbc.edu}

\author{Mansa Krishna}
\affiliation{
\institution{Department of Earth Sciences, Dartmouth College}
\country{USA}
}
\email{mansa.krishna.gr@dartmouth.edu}

\author{Homayra Alam}
\affiliation{
\institution{University of Maryland, Baltimore County (UMBC)}
\country{USA}
}
\email{halam3@umbc.edu}

\author{Mostafa Cham}
\affiliation{
\institution{University of Maryland, Baltimore County (UMBC)}
\country{USA}
}
\email{mcham2@umbc.edu}

\author{Omar Faruque}
\affiliation{
\institution{University of Maryland, Baltimore County (UMBC)}
\country{USA}
}
\email{omarf1@umbc.edu}

\author{Gong Cheng}
\affiliation{
\institution{Department of Earth Sciences, Dartmouth College}
\country{USA}
}
\email{gong.cheng@dartmouth.edu}

\author{Jianwu Wang}
\affiliation{
\institution{University of Maryland, Baltimore County (UMBC)}
\country{USA}
}
\email{jianwu@umbc.edu}

\author{Mathieu Morlighem}
\affiliation{
\institution{Department of Earth Sciences, Dartmouth College}
\country{USA}
}
\email{mathieu.morlighem@dartmouth.edu}

\author{Vandana Janeja}
\affiliation{
\institution{University of Maryland, Baltimore County (UMBC)}
\country{USA}
}
\email{vjaneja@umbc.edu}

\renewcommand{\shortauthors}{Tama et al.}

\begin{document}

\begin{abstract}
Mapping Greenland's subglacial topography is critical for projecting the future mass loss of the ice sheet and its contribution to global sea-level rise. However, the complex and sparse nature of observational data, particularly information about the bed topography under the ice sheet, significantly increases the uncertainty in model projections. Bed topography is traditionally measured by airborne ice-penetrating radars that measure the ice thickness directly underneath the aircraft, leaving data gaps of tens of kilometers in between flight lines. This study introduces a deep learning framework, \textbf{DeepTopoNet}, that integrates radar-derived ice thickness observations and BedMachine Greenland data through a novel dynamic loss-balancing mechanism. Among all efforts to reconstruct bed topography, BedMachine has emerged as one of the most widely used datasets, combining mass conservation principles and ice thickness measurements to generate high-resolution bed elevation estimates. The proposed loss function adaptively adjusts the weighting between radar and BedMachine's bed, ensuring robustness in areas with limited radar coverage while leveraging the high spatial resolution of BedMachine's bed estimates. Our approach incorporates gradient-based and trend surface features to enhance model performance and utilizes a convolutional neural network (CNN) architecture (i.e., \textbf{BedTopoCNN}) designed for subgrid-scale predictions. By systematically testing on the Upernavik Isstr{\o}m) region in West Greenland, the model achieves high accuracy (MAE: 12.49 m, RMSE: 19.38 m, and R$^{2}$: 0.99), outperforming baseline methods in reconstructing subglacial terrain. This work demonstrates the potential of deep learning in bridging observational gaps, providing a scalable and efficient solution to inferring subglacial topography. This framework  paves the way for improved predictions of ice sheet flow and sea level rise.
\end{abstract}



\maketitle

\section{Introduction}
The Greenland Ice Sheet is a major contributor to global sea-level rise, driven by warmer atmospheric temperatures and the intrusion of warm ocean currents in fjords. Rising sea level poses a critical threat to coastal communities worldwide, increasing the risk of flooding, shoreline erosion, and habitat loss~\cite{ipcc2023}. Estimating its contribution to sea level requires a detailed understanding of subglacial topography, which influences ice flow and basal friction~\cite{barnes2021transferability,aakesson2021future}. However, accurately inferring the bed topography remains a significant challenge due to limited observational coverage, especially in the ice sheet interior. Traditional mapping methods, such as mass conservation models and geostatistic techniques~\cite{morlighem2011mass,bamber2001new,bamber2013new}, have been instrumental but often struggle in regions of slower flow.

Deep learning~\cite{lecun2015deep}, with its ability to learn complex relationships from multimodal data, has shown promise in solving geophysical problems. In this study, we present a deep learning framework for inferring Greenland’s subglacial topography, integrating radar-derived observations and BedMachine~\cite{morlighem2017bedmachine,morlighem2014deeply} data. A key innovation in our approach is the introduction of a dynamic loss-balancing mechanism that adaptively adjusts the contribution of radar and BedMachine data to the training process. This ensures robust performance in regions with sparse radar data while leveraging the high-resolution predictive capabilities of BedMachine.

Our methodology involves a convolutional neural network (CNN) architecture~\cite{lecun2015deep} designed to capture spatial patterns in surface and subglacial features. By incorporating gradient-based covariates and trend-surface features, the model enhances its capacity to learn subgrid-scale variations and reduces the uncertainty in bed topography prediction. Additionally, the dynamic loss function ensures that the model optimally balances data-driven predictions and physical constraints derived from multiple data sources.

In this paper, we evaluate our proposed framework using a subset of Greenland’s bed topography. Our results demonstrate the model's ability to accurately reconstruct subglacial terrain, outperforming traditional interpolation methods and baseline machine learning approaches. This work highlights the potential of deep learning in addressing critical geophysical challenges and provides a scalable solution for improving ice sheet models. The insights gained from this study contribute to a deeper understanding of subglacial dynamics and their implications for sea-level rise projections.

To summarize, the contributions of this paper are threefold: (i) \textbf{DeepTopoNet Framework}: A novel deep learning framework integrating radar observations and BedMachine data, featuring a CNN architecture enhanced with gradient-based covariates and trend-surface features; (ii) \textbf{Dynamic Loss-Balancing Mechanism}: An adaptive mechanism ensuring robust performance in regions with sparse radar data by balancing contributions from multimodal data sources; and (iii) \textbf{Comprehensive Evaluation}: Extensive benchmarking across Greenland's Upernavik Isstr{\o}m sector and unseen regions, demonstrating superior performance over traditional interpolation and machine learning methods.

\section{Related Work}
\textbf{Spatial Interpolation}. Spatial interpolation~\cite{li2000comparison,lam1983spatial} is a crucial methodology for estimating values at unobserved locations based on spatial data and has wide applications in hydrology, meteorology, and geosciences. Traditional interpolation techniques include deterministic methods, such as Inverse Distance Weighting (IDW)~\cite{li2008review} and Triangulated Irregular Network (TIN)~\cite{feng2024critical}, which rely on predefined spatial correlation functions. Geostatistical methods~\cite{wackernagel2003ordinary,wackernagel2003universal}, like Kriging and its variants (e.g., Ordinary Kriging and Universal Kriging), use spatial statistics and variogram models to infer underlying spatial patterns~\cite{bamber2013new}. However, these methods are often limited by their reliance on fixed assumptions about spatial correlations and Gaussian process assumptions, which may not hold for many real-world datasets. Moreover, they rely solely on bed topography observations, which are sparse, and do not utilize any other ancillary dataset which may help guide the interpolation (e.g., deep trough generally coincide with observed high ice flow speed).

Recent developments in machine learning have enabled more sophisticated approaches to spatial interpolation. For example, Kriging Convolutional Networks (KCN)~\cite{appleby2020kriging} integrate the strengths of graph neural networks (GNNs)~\cite{kipf1609semi} and kriging to enhance spatial data modeling. KCN leverages GNN-based propagation mechanisms while maintaining kriging-like inductive capabilities, offering flexibility and improved accuracy over standard kriging techniques. Similarly, the Graph for Spatial Interpolation (GSI) model adapts GNN architectures to dynamically learn spatial correlations while addressing limitations of fixed adjacency matrices, demonstrating effectiveness for rainfall spatial interpolation~\cite{li2023rainfall}. These approaches underscore the growing role of machine learning in addressing challenges such as irregular spatial patterns and dynamic spatial dependencies.

\textbf{Subglacial Bed Topography Prediction}. Inferring the subglacial bed topography has been a critical focus in glaciology due to its implications for ice sheet mass balance and its contribution to global sea-level rise. Recent advancements have integrated machine learning, physics-based models, and hybrid approaches to tackle this complex problem. Cheng et al. (2024)~\cite{cheng2024forward} applied Physics-Informed Neural Networks (PINNs) to model ice flow dynamics for Helheim Glacier in Greenland. The study demonstrated PINNs’ capability to infer basal friction coefficients and interpolate sparse ice thickness observations with high accuracy. The framework uniquely incorporates the conservation of momentum (i.e., the shallow-shelf/shelfy-stream approximation)~\cite{macayeal1989large,morland1987unconfined} as soft constraints in the loss function, improving predictions where data are sparse and enhancing the versatility of the model. Yi et al. (2023)~\cite{yi2023evaluating} evaluated nine machine learning and statistical models, including XGBoost, Gaussian Process Regression, and hybrid models, to predict Greenland’s subglacial bed topography. The study highlighted the strong performance of XGBoost when paired with kriging-based interpolation, achieving the best RMSE and R$^{2}$.  Similarly, Leong and Horgan (2020)~\cite{leong2020deepbedmap} introduced DeepBedMap, a deep learning framework based on generative adversarial networks, to produce high-resolution bed elevation maps of Antarctica. Their approach combined multi-resolution spatial data, such as surface elevation and ice velocity, to improve the representation of subglacial terrain roughness. These studies underscore the importance of combining domain knowledge with advanced computational tools to address challenges in regions with sparse observations. This paper builds on these foundations, proposing a robust framework that integrates multimodal data, deep learning, and loss balancing to further enhance predictive performance.

\section{Preliminary}
Accurately inferring Greenland's subglacial topography is essential for understanding ice dynamics and predicting sea-level rise. However, this task is challenging due to sparse radar observations, limited direct measurements, and the inherent complexity of the ice-bed interactions. The goal of this paper is to develop a predictive framework that leverages surface observations and physical constraints to estimate bed topography, integrating multiple data sources. Let $b(x,y)$ be the bed elevation at coordinates $(x,y)$. Given sparse radar measurements ${r}(x,y)$ and the modeled topography data from BedMachine ${m}(x,y)$, the objective is to predict $b(x,y)$ with high spatial resolution across the domain $\Omega$. This can be formulated as: 
\begin{equation} 
b(x,y) = f\left({s}(x,y),\textbf{v}(x,y),\frac{\partial {h}}{\partial t}(x,y),\mbox{SMB}(x,y),\textbf{g}(\nabla \textbf{x}), \textbf{t}(x,y)\right), 
\end{equation}
where ${s}(x,y)$ is surface elevation, $\textbf{v}(x,y) = ({v}_{x}(x,y),{v}_{y}(x,y))$ is surface ice velocity components, $\frac{\partial {h}}{\partial t}(x,y)$ is surface thickening rate over time, $\mbox{SMB}(x,y)$ is surface mass balance (e.g., snow accumulation or surface melt), $\nabla \textbf{x}$ is gradients of all input features (e.g., surface elevation, surface velocity, surface thickening rate, and SMB), and $\textbf{t}(x,y)$ is trend surface features generated from polynomial fitting of spatial coordinates.

\section{Proposed Framework}
\begin{figure*}[ht!]
    \centering
    \includegraphics[width=1\linewidth]{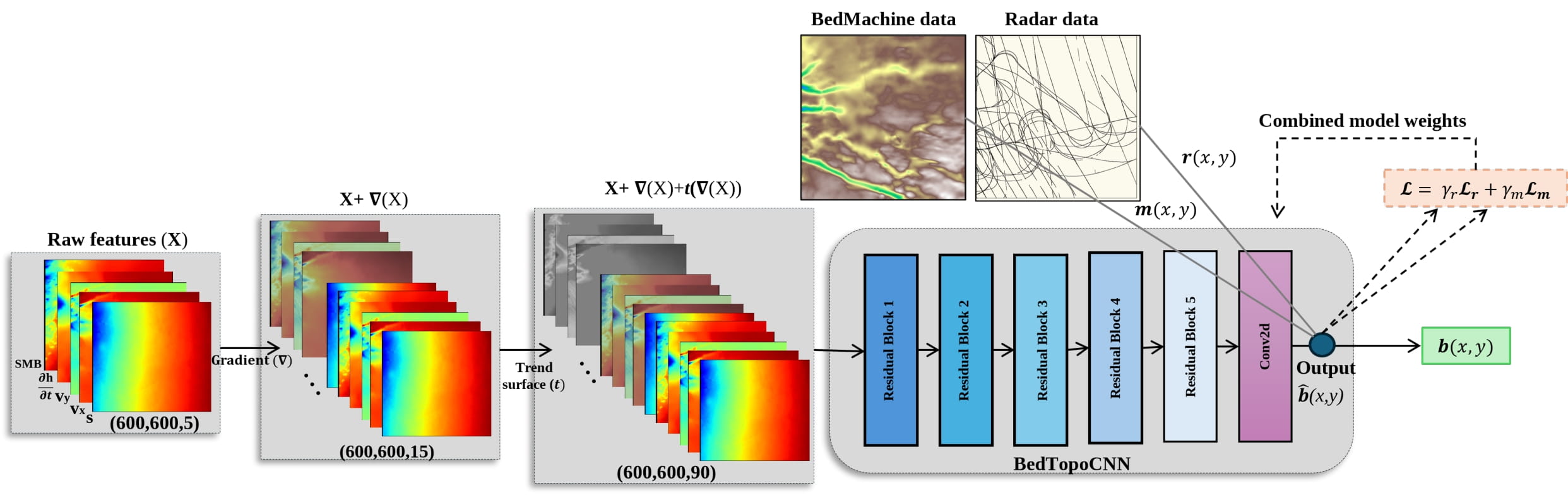}
    \caption{Overview of the proposed DeepTopoNet framework, including feature extraction, BedTopoCNN architecture, and dynamic loss balancing.}
    \label{framework}
\end{figure*}

The proposed framework, \textbf{DeepTopoNet}, integrates radar and BedMachine data to infer Greenland's subglacial topography. Gradient and trend surface features enhance spatial variability capture, and the data is processed through \textbf{BedTopoCNN}, a convolutional neural network with residual blocks. A dynamic loss function balances contributions from radar and BedMachine data for robust predictions. The overall framework is shown in Figure~\ref{framework}.

\subsection{Prior Physical Model (BedMachine)}
BedMachine~\cite{morlighem2017bedmachine,morlighem2020deep} is the most widely used subglacial topography map of Greenland. It employs a mass conservation approach in fast flowing regions, combining radar-derived ice thickness measurements, satellite velocity data, and surface mass balance reconstructions to produce a high-resolution bed topography map. The idea of mass conservation is to infer the bed elevation by ensuring conservation of mass: $\nabla\cdot(s-b){\bf v} = \dot{a}$, where  $\dot{a}$ is the apparent mass balance that combines observed thickness change and surface mass balance from a regional climate model. BedMachine ensures physically realistic reconstructions in regions of fast-flowing ice, but remains highly uncertain in slower moving regions where it is based on a simple interpolation scheme that connects existing radar data. The output is a seamless topographic map with horizontal resolution up to 150 m. In this study, BedMachine data serves as a reference for the target variable, subglacial bed topography ($b(x,y)$). The proposed DeepTopoNet framework goes beyond the interpolation based approach and the reliance on mass conservation. 

\subsection{Input Features and Covariates}
To enhance predictive accuracy, the model incorporates both raw data and derived features: (i) \textbf{Gradient Features}~\cite{jacobs2005image}. The gradient of raw input features \textbf{X} is computed as partial derivatives along $x$- and $y$-directions: $\nabla\textbf{X}=\left(\frac{\partial\textbf{X}}{\partial x},\frac{\partial\textbf{X}}{\partial y} \right)$, where $\textbf{X}\in\{{s}(x,y),{v}_{x}(x,y)$, ${v}_{y}(x,y), \frac{\partial {h}}{\partial t}(x,y),\mbox{SMB}(x,y)\}$. These gradients provide additional spatial information on the variability of each feature and are concatenated to the input feature set. (ii) \textbf{Trend Surface Features}~\cite{agterberg2023trend}. For each input feature \textbf{X}, trend surface features are derived by fitting a polynomial of degree $d=2$ over the spatial coordinates: $\textbf{t}_{\textbf{X}}(x,y)=\sum_{i+j\leq d} a_{ij}x^{i}y^{j}$, where $a_{ij}$ are coefficients estimated for \textbf{X}. These trend surfaces capture global spatial patterns and improve the model's ability to generalize in regions with sparse data.

The integration of gradient features and trend surface features into DeepTopoNet represents a novel approach for subglacial topography estimation. Gradient features enhance local spatial sensitivity by capturing fine-scale variations in ${s}(x,y)$, $\textbf{v}(x,y)$, $\frac{\partial {h}}{\partial t}(x,y)$, and $\mbox{SMB}(x,y)$, while trend surface features provide a global spatial context by modeling large-scale variations using polynomial fitting. While these techniques have been used individually in geostatistics~\cite{briggs1974machine,hutchinson1995interpolating,wang2023unraveling,merriam1966geologic} and image processing~\cite{riba2020kornia,canny1986computational,sobel1970camera,soria2023tiny}, their combined use in a deep learning framework for sparse spatial predictions is unprecedented. Our hybrid approach improves the model’s ability to generalize in regions with sparse observations, bridging local feature variability with global spatial trends to enhance topographic reconstruction accuracy.

\subsection{Model Architecture}
\begin{figure}[ht!]
    \centering
    \includegraphics[width=0.9\linewidth]{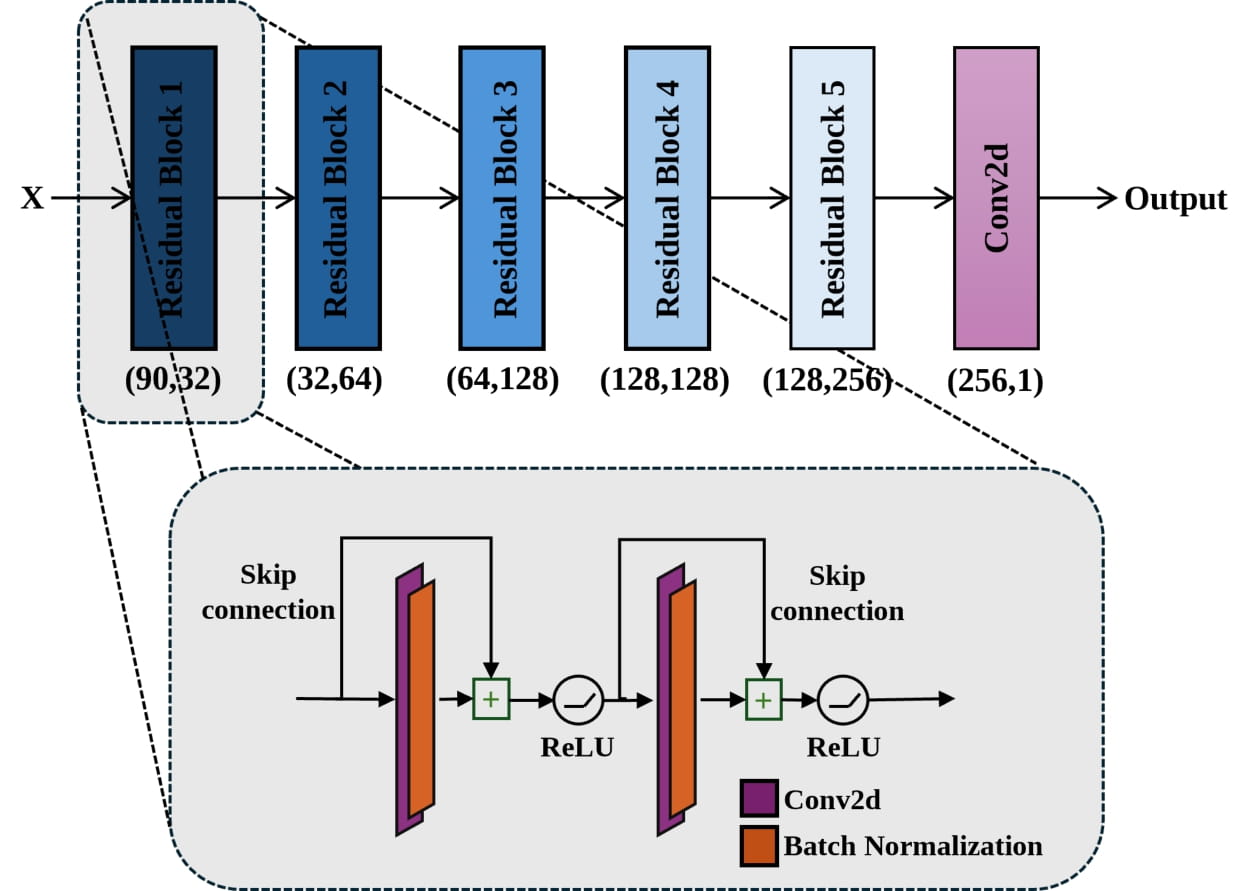}
    \caption{Architecture of the proposed BedTopoCNN model.}
    \label{model_architecture}
\end{figure}

The proposed model architecture, \textbf{BedTopoCNN}, shown in Figure~\ref{model_architecture} is a deep convolutional neural network specifically designed for subglacial topography prediction. Building on residual learning principles introduced by He et al. (2016)~\cite{he2016deep}, the architecture employs a series of \textbf{Residual Blocks (RBs)} to capture spatial features from multimodal input data effectively while addressing challenges such as vanishing gradients during training. Each \textbf{RB} consists of two convolutional layers with a kernel size of $3\times3$, padding, and batch normalization to ensure feature stability. Non-linear transformations are applied using \textbf{ReLU activation}, while a \textbf{skip connection} (adapted from~\cite{he2016deep}) allows the input to bypass the convolutional layers, preserving features and mitigating gradient vanishing issues. Additionally, \textbf{Dropout regularization} is incorporated within the blocks to prevent overfitting, which is particularly important in regions with sparse radar data~\cite{cai2025siamese,leong2020deepbedmap}.

The overall architecture begins with an \textbf{Input Layer} that processes the preprocessed input features. This is followed by \textbf{five RBs}, where the number of filters progressively increases from 32 to 256 to extract hierarchical spatial features. Finally, the network concludes with a $1\times1$ convolutional layer that reduces the output to a single channel, corresponding to the predicted bed topography. The combination of RBs and the hierarchical design allows the network to effectively model complex relationships between input features and the target subglacial topography. 

\subsection{Loss Function}
We propose a \textbf{dynamic loss function} to guide the training process by balancing the contributions of radar data ${r}(x,y)$ and BedMachine data ${m}(x,y)$. While dynamic loss balancing has been explored in multi-task learning~\cite{kendall2018multi} and semi-supervised settings~\cite{wang2021self}, our approach uniquely adapts to geospatial data sparsity patterns rather than task uncertainty or pseudo-label confidence. Furthermore, in contrast to PINNs, where loss weights are often manually tuned or fixed~\cite{raissi2019physics}, our method dynamically adjusts the weighting terms $\upgamma_{r}$ and $\upgamma_{m}$ based on the local density of radar observations in each batch. The total loss $\mathcal{L}$ is expressed as $\mathcal{L}=\upgamma_{r}\mathcal{L}_{r}+\upgamma_{m}\mathcal{L}_{m}$, where:
\begin{itemize}
    \item $\mathcal{L}_{r}=\frac{1}{|\Omega_{r}|}\sum_{(x,y)\in\Omega_{r}}(\hat{b}(x,y)-{r}(x,y))^{2}$ is the radar loss, penalizing discrepancies in radar-covered regions $\Omega_{r}$, 
    \item $\mathcal{L}_{m}=\frac{1}{|\Omega_{m}|}\sum_{(x,y)\in\Omega_{m}}(\hat{b}(x,y)-{m}(x,y))^{2}$ is the BedMachine loss, evaluating prediction accuracy in non-radar regions $\Omega_{m}$, and $\upgamma_{r}=\frac{\mathcal{L}_{m}}{\mathcal{L}_{r}+\mathcal{L}_{m}+\epsilon}$,
    \item $\upgamma_{m}=\frac{\mathcal{L}_{r}}{\mathcal{L}_{r}+\mathcal{L}_{m}+\epsilon}$ are dynamic weights that adjust during training, ensuring balanced optimization with $\epsilon$ as a small smoothing factor to avoid division by zero.
\end{itemize}
The magnitudes of $\mathcal{L}_{r}$ and $\mathcal{L}_{m}$ are inherently influenced by the proportion of radar-covered ($\Omega_{r}$) and non-radar ($\Omega_{m}$) pixels in each mini-batch. In radar-dense batches, $\mathcal{L}_{r}$ is computed over many radar points and tends to dominate, resulting in a smaller $\upgamma_{r}$ and a larger $\upgamma_{m}$. In radar-sparse batches, $\mathcal{L}_{m}$ dominates, increasing $\upgamma_{r}$ and reducing $\upgamma_{m}$. This coupling between loss magnitude and data coverage means that the weighting terms implicitly adapt to the \textit{local radar density} of each batch without requiring explicit density computation or manual tuning. 

The objective is to train a a deep convolutional neural network model $f_{\theta}$ with parameters $\theta$ to minimize the total loss $\mathcal{L}$, ensuring robust predictions across radar-sparse and radar-dense regions: $\hat{\theta}=\arg\min_{\theta}\mathcal{L}(\theta)$.

\section{Experiments}
In this section, we evaluate the accuracy of~\textbf{DeepTopoNet} and demonstrate its predictive capability across the entire grid map.

\subsection{Experimental Setup}
\subsubsection{Dataset}
The dataset used in this study is obtained from the Upernavik Isstr{\o}m region in Central West Greenland, encompassing four distinct sub-regions. Each sub-region covers a spatial extent of $600\times600$ grid cells, representing an area with diverse ice dynamics and varying data density. The number of radar data points differs across the sub-regions, reflecting the uneven distribution of observations. The dataset includes several observed variables critical for subglacial topography inference:
\begin{itemize}
    \item ${s}(x,y)$: Surface elevation~\cite{howat2014greenland}, representing the height of the ice sheet surface above a reference level.
    \item $\textbf{v}(x,y) = ({v}_{x}(x,y),{v}_{y}(x,y))$: Surface ice velocity components~\cite{joughin2010greenland} in the $x$- and $y$-directions, describing ice flow dynamics.
    \item $\frac{\partial {h}}{\partial t}(x,y)$: Surface thickening or thinning rate~\cite{millan2022ice}, indicating changes in ice sheet thickness over time.
    \item $\mbox{SMB}(x,y)$: Surface mass balance~\cite{imbie2020mass,noel2018modelling}, which accounts for net snow accumulation or surface melt rates.
\end{itemize}
These variables provide essential input features for the model, capturing spatial and temporal variations in surface and subsurface ice dynamics. The dataset's diverse sub-regions allow the DeepTopoNet to generalize across areas with different radar data availability and ice sheet characteristics, enhancing the robustness of subglacial topography predictions.

\begin{figure}[ht!]
    \centering
    \includegraphics[width=0.45\textwidth]{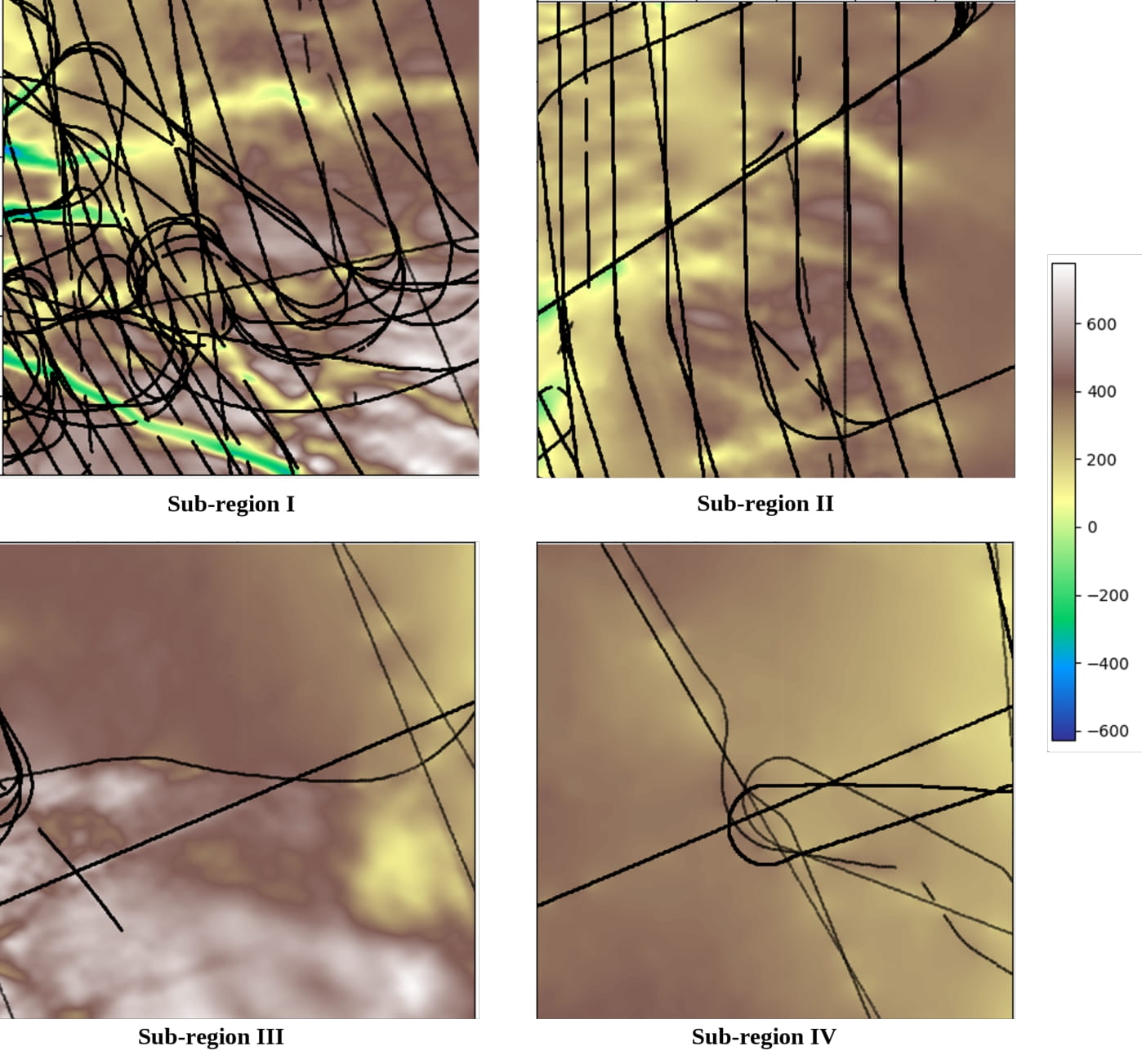}
    \caption{Visualization of BedMachine topography with radar observation points overlaid for each sub-region in the Upernavik Isstr{\o}m region. The figure highlights the spatial distribution of radar data relative to the underlying bed topography.}
    \label{radar_data_bm}
\end{figure}

The spatial relationship between radar observations and BedMachine topography data is critical for understanding the integration of data in subglacial bed topography modeling. A detailed visualization of the BedMachine data with overlaid radar observation points for each sub-region in the Upernavik Isstr{\o}m region is provided in Figure~\ref{radar_data_bm}. This figure highlights the spatial distribution of radar data relative to the underlying bed topography, showcasing the coverage of radar tracks and emphasizing the importance of filling data gaps in regions with sparse radar measurements.

\begin{table}[ht!]
\caption{Summary of the dataset characteristics for the Upernavik Isstr{\o}m , including bed height range and radar data point distribution.}
\label{dataset_summarization}
\centering
\resizebox{0.47\textwidth}{!}{%
\begin{tabular}{cccc}
\hline
Sub-regions&Min\_bed\_height&Max\_bed\_height&\#Radar data points\\
\hline
I&-478.7407&778.9677&316,559\\
II&-144.0005&646.0179&238,899\\
III&115.8914&762.7296&19,509\\
IV&106.4424&428.6476&57,340\\
\hline
\end{tabular} 
}
\end{table}

\subsubsection{Evaluation Metrics}
The model's performance is assessed during training and validation, as well as through a comprehensive evaluation on the full prediction grid. For the evaluation of full grid predictions, several standard metrics are employed to comprehensively measure the model's accuracy and structural fidelity such as MAE, RMSE, Structural Similarity Index Measure (SSIM)~\cite{wang2004image}, and Peak Signal-to-Noise Ratio (PSNR)~\cite{hore2010image}.
These metrics, commonly used in spatial modeling and image reconstruction tasks, provide a holistic evaluation of the model's accuracy, structural preservation, and robustness for full grid predictions.

Moreover, a domain-specific metric, the \textbf{Terrain Ruggedness Index (TRI)}~\cite{reily1999terrain}, is used as a quantitative measure to evaluate local variations in terrain elevation by capturing the differences between a cell and its neighboring elevation values. TRI provides insight into how well a model preserves the spatial variability and ruggedness of the terrain, which is critical for applications such as glaciological modeling and subglacial flow analysis. By comparing TRI values between the predicted topography and the reference BedMachine data, we assess the model's ability to reproduce terrain features accurately. For a given digital elevation model (DEM), TRI is calculated for each grid cell by measuring the elevation differences between the cell and its neighboring cells. The value represents the root mean square of these differences, capturing the local terrain variability. To evaluate the model predictions, we leverage relative difference (\%), which is the relative difference that quantifies how closely the ruggedness of the predicted topography aligns with the reference BedMachine data:
\begin{equation}
    \text{Relative Difference (\%)} = \frac{|TRI_{predicted}-TRI_{BedMachine}|}{TRI_{predicted}}\times 100
\end{equation}
where a lower relative difference indicates better alignment between the predicted and reference ruggedness.

Algorithm~\ref{tri_calculation} describes the calculation of TRI for a grid-based Digital Elevation Model (DEM). For each cell in the $M\times N$ elevation grid, the algorithm calculates the squared differences in elevation between the cell and its neighboring cells (defined by offsets $\mathcal{N}$). These differences are accumulated across all grid cells and normalized by the total number of neighboring pairs. The TRI is then computed as the square root of the mean squared elevation differences, providing a measure of terrain variability. This approach captures fine-scale topographic details, making it useful for characterizing subglacial landscapes.  

\begin{algorithm}[]
\caption{Terrain Ruggedness Index (TRI) Calculation}
\label{tri_calculation}
\SetKwInOut{Input}{Input}
\SetKwInOut{Output}{Output}
\Input{2D grid of elevation values $\mathbf{E}$ with dimensions $M \times N$.}
\Output{TRI value for the grid.}
\SetAlgoLined
\DontPrintSemicolon
\textbf{Initialize:} $\Delta h \gets 0$ (sum of squared elevation differences) \;
Define neighbor offsets $\mathcal{N} \gets \{(0, 1), (1, 0), (1, 1), (1, -1)\}$ \;
\For{$i \gets 0$ \textbf{to} $M-1$}{
    \For{$j \gets 0$ \textbf{to} $N-1$}{
        \ForEach{$(di, dj) \in \mathcal{N}$}{
            Compute elevation difference: $\Delta h_{i,j} \gets (\mathbf{E}[i, j] - \mathbf{E}[i+di, j+dj])^2$ \;
            Accumulate: $\Delta h \gets \Delta h + \Delta h_{i,j}$ \;
        }
    }
}
Normalize by total pairs: $N_{\text{pairs}} \gets (M-1) \cdot (N-1) \cdot |\mathcal{N}|$ \;
\textbf{Compute TRI:} $\text{TRI} \gets \sqrt{\frac{\Delta h}{N_{\text{pairs}}}}$ \;
\Return $\text{TRI}$
\end{algorithm}

\begin{table*}[ht!]
\caption{Performance comparison of \textbf{DeepTopoNet} against baseline methods across four sub-regions for bed topography prediction. The best-performing results are highlighted in \textbf{bold}, while the second-best results are underlined.}
\label{result_quantitative}
\centering
\resizebox{1\textwidth}{!}{
\begin{tabular}{l|lllll|lllll|lllll|lllll}
\hline
\multirow{2}{*}{Method}&\multicolumn{5}{|c|}{Sub-region I}&\multicolumn{5}{|c|}{Sub-region II}&\multicolumn{5}{|c}{Sub-region III}&\multicolumn{5}{|c}{Sub-region IV}\\
\cline{2-21}
&MAE$\downarrow$&RMSE$\downarrow$&R$^{2}\uparrow$&SSIM$\uparrow$&PSNR$\uparrow$&MAE$\downarrow$&RMSE$\downarrow$&R$^{2}\uparrow$&SSIM$\uparrow$&PSNR$\uparrow$&MAE$\downarrow$&RMSE$\downarrow$&R$^{2}\uparrow$&SSIM$\uparrow$&PSNR$\uparrow$&MAE$\downarrow$&RMSE$\downarrow$&R$^{2}\uparrow$&SSIM$\uparrow$&PSNR$\uparrow$\\
\hline\hline
IDW~\cite{johnston2001using}&63.15&101.05&0.65&0.78&21.90&29.90&41.84&0.82&\underline{0.87}&25.52&52.16&74.92&0.58&\textbf{0.91}&18.72&22.99&34.31&0.67&0.94&19.46\\
RBF~\cite{heryudono2010radial}&65.95&99.93&0.66&0.77&22.00&35.28&53.38&0.70&0.85&23.40&57.48&76.52&0.56&0.90&18.54&20.56&26.72&0.80&0.94&21.63\\
RF~\cite{sekulic2020random}&68.00&99.87&0.66&0.69&22.00&79.90&97.13&0.01&0.59&18.21&61.90&93.35&0.34&0.84&16.81&24.46&34.13&0.67&0.84&19.50\\
MLP-PE~\cite{li2021learnable}&113.17&195.27&-0.30&0.48&16.18&112.51&170.17&-2.03&0.28&13.34&145.43&238.89&-3.30&0.41&8.65&88.49&151.62&-5.46&0.19&6.55\\
VAE+XGBoost~\cite{yi2023evaluating}&78.12&108.54&0.60&0.75&21.28&42.52&59.79&0.63&0.75&22.42&57.91&85.77&0.45&0.85&17.55&22.95&30.88&0.73&0.83&20.37\\
U-Net~\cite{ronneberger2015u}&56.43&75.59&0.81&0.82&24.42&46.81&68.57&0.51&0.72&21.23&69.83&106.39&0.15&0.64&15.68&14.17&20.91&0.88&0.88&23.76\\
U-Net++~\cite{zhou2018unet++}&88.36&114.47&0.55&0.78&20.82&41.83&65.24&0.55&0.76&21.66&61.42&94.61&0.33&0.68&16.70&9.62&14.26&0.94&0.88&27.08\\
U-Net3+~\cite{huang2020unet}&51.33&70.92&0.83&0.77&24.98&33.78&47.82&0.76&0.80&24.36&62.94&97.77&0.28&0.72&16.41&24.95&34.99&0.66&0.88&19.28\\
Att. U-Net~\cite{oktay2018attention}&\underline{21.88}&\underline{33.33}&\underline{0.96}&\underline{0.90}&\underline{31.53}&\underline{21.35}&\underline{33.21}&\underline{0.89}&\underline{0.87}&\underline{27.53}&\textbf{20.65}&\textbf{31.54}&\textbf{0.93}&\underline{0.90}&\textbf{26.24}&\underline{7.33}&\underline{11.75}&\underline{0.96}&\underline{0.94}&\underline{28.76}\\
\textbf{DeepTopoNet}&\textbf{12.49}&\textbf{19.38}&\textbf{0.99}&\textbf{0.94}&\textbf{36.25}&\textbf{18.26}&\textbf{28.90}&\textbf{0.91}&\textbf{0.88}&\textbf{28.73}&\underline{21.00}&\underline{32.69}&\underline{0.92}&\textbf{0.91}&\underline{25.93}&\textbf{4.74}&\textbf{7.65}&\textbf{0.98}&\textbf{0.96}&\textbf{32.49}\\
\hline
\end{tabular}
}
\end{table*}

\begin{table*}[ht!]
\caption{Results of the ablation study, evaluating the impact of removing stride, gradient features ($\nabla$), trend surface ($\textbf{t}$) features, and BedMachine loss ($\mathcal{L}_{m}$) on the performance of DeepTopoNet across four sub-regions. Best results are highlighted in bold.}
\label{ablation_result}
\centering
\resizebox{1\textwidth}{!}{
\begin{tabular}{l|lllll|lllll|lllll|lllll}
\hline
\multirow{2}{*}{Method}&\multicolumn{5}{|c|}{Sub-region I}&\multicolumn{5}{|c|}{Sub-region II}&\multicolumn{5}{|c}{Sub-region III}&\multicolumn{5}{|c}{Sub-region IV}\\
\cline{2-21}
&MAE$\downarrow$&RMSE$\downarrow$&R$^{2}\uparrow$&SSIM$\uparrow$&PSNR$\uparrow$&MAE$\downarrow$&RMSE$\downarrow$&R$^{2}\uparrow$&SSIM$\uparrow$&PSNR$\uparrow$&MAE$\downarrow$&RMSE$\downarrow$&R$^{2}\uparrow$&SSIM$\uparrow$&PSNR$\uparrow$&MAE$\downarrow$&RMSE$\downarrow$&R$^{2}\uparrow$&SSIM$\uparrow$&PSNR$\uparrow$\\
\hline\hline
w/o stride,$\nabla$, \& $\textbf{t}$&42.84&62.66&0.87&0.80&26.05&28.91&41.00&0.82&0.85&25.70&22.44&34.52&0.91&\textbf{0.91}&25.46&87.99&457.06&-57.74&0.88&-3.04\\ 
w/o stride, w/ $\nabla$ \& $\textbf{t}$&23.27&36.26& 0.96&0.88&30.80&18.42&\textbf{28.78}&\textbf{0.91}&\textbf{0.89}&\textbf{28.77}&\textbf{17.37}&\textbf{27.55}&\textbf{0.94}&0.90&\textbf{27.41}&4.80&\textbf{7.53}&\textbf{0.98}&\textbf{0.96}&\textbf{32.63}\\
w/o $\mathcal{L}_{m}$&39.21&67.38&0.85&0.79&25.42&34.44&54.09&0.69&0.73&23.29&65.13&104.42&0.18&0.66&15.84&92.42&106.00&-2.16&0.61&9.66\\
\textbf{DeepTopoNet}&\textbf{12.49}&\textbf{19.38}&\textbf{0.99}&\textbf{0.94}&\textbf{36.25}&\textbf{18.26}&28.90&\textbf{0.91}&0.88&28.73&21.00&32.69&0.92&\textbf{0.91}&25.93&\textbf{4.74}&7.65&\textbf{0.98}&\textbf{0.96}&32.49\\ 
\hline
\end{tabular}
}
\end{table*}

\subsubsection{Baselines}
We compare the performance of the proposed DeepTopoNet against three categories of bed topography prediction baselines, which include traditional interpolation methods: IDW~\cite{johnston2001using} and RBF~\cite{heryudono2010radial}; pixel-based machine learning methods: random forest (RF)~\cite{sekulic2020random}, MLP with positional encoding (MLP-PE)~\cite{li2021learnable}, and VAE with XGBoost~\cite{yi2023evaluating}; and patch-based deep learning models: U-Net~\cite{ronneberger2015u}, U-Net++~\cite{zhou2018unet++}, U-Net3+~\cite{huang2020unet}, and Attention U-Net~\cite{oktay2018attention}. Below is a brief explanation of each method:
\begin{itemize}
    \item IDW~\cite{johnston2001using}: A deterministic interpolation technique that assigns weights to nearby points inversely proportional to their distance from the target point. This method emphasizes closer points for prediction, assuming spatially smooth variations.
    \item RBF~\cite{heryudono2010radial}: An interpolation method that uses radial basis functions to create a smooth surface fitting the data. RBF is known for its flexibility in capturing localized spatial variations.
    \item RF~\cite{sekulic2020random}: This baseline utilizes an RF regressor with additional spatial enhancements to predict bed topography.
    \item MLP with positional encoding (MLP-PE)~\cite{li2021learnable}: This baseline employs a Multilayer Perceptron (MLP) model enhanced with sinusoidal positional encodings to incorporate spatial context into the predictions. 
    \item VAE with XGBoost~\cite{yi2023evaluating}: This baseline leverages a Variational Autoencoder (VAE) to encode spatial patterns into a latent representation, followed by an XGBoost regressor to predict subglacial topography with enhanced feature extraction. 
    \item U-Net~\cite{ronneberger2015u}: A widely used convolutional neural network for segmentation tasks, featuring an encoder-decoder structure with skip connections to preserve spatial details.
    \item U-Net++~\cite{zhou2018unet++}: An extension of U-Net that introduces dense skip connections to improve feature aggregation and capture finer details.
    \item U-Net3+~\cite{huang2020unet}: A further enhancement of U-Net with nested and multi-scale skip connections, designed to leverage information across different feature resolutions.
    \item Attention U-Net~\cite{oktay2018attention}: An advanced U-Net variant that incorporates attention mechanisms to focus on critical regions of the input, enhancing the model’s ability to capture important spatial features. 
\end{itemize}

\subsubsection{Experimental Settings}
During training, the dataset is split into 80\% for training and 20\% for validation, ensuring an unbiased evaluation of the model's generalization capabilities. The training data is processed in overlapping patches of size $16\times16$ with a stride of 8, increasing the effective dataset size and capturing both local and contextual spatial features. The model is trained with a batch size of 16 using the Adam optimizer for 20,000 epochs. A cyclic learning rate scheduler dynamically adjusts the learning rate during training. The loss function $\mathcal{L}$ dynamically balances radar and BedMachine data contributions. Early stopping is employed with a patience threshold of 5,000 epochs. The best model is saved when the validation loss improves, ensuring optimal performance and preventing overfitting. We present hyperparameter settings for each method as follows.
\begin{itemize}
    \item IDW. Number of Nearest Neighbors ($k$): 4000, power parameter ($p$): 2, distance threshold = 10$^{10}$.
    \item RBF. The $multiquadratic$ function is used to capture smooth variations in the data, smoothness parameter ($\epsilon$): 2, small regularization term: $1\times10^{-3}$. 
    \item Random forest. The hyperparameter settings were optimized using Optuna~\cite{akiba2019optuna}, a framework for automated hyperparameter optimization. These hyperparameters were tuned as follows. $n\_estimators$: 100$\sim$1000, $min\_samples\_split$: 2$\sim$10, $min\_samples\_leaf$: 1$\sim$10, $max\_features$: $sqrt$. The objective function optimized the RMSE on the test set. The optimization process was conducted using 100 trials with the TPE (Tree-structured Parzen Estimator) sampler for efficient exploration of the search space. 
    \item MLP with Positional Encoding (MLP-PE). Sinusoidal positional encodings are added to the input data to incorporate spatial context, with $num\_encodings$: 15. The input features are augmented using gradient covariates, polynomial trend surface, and nearest neighbor values using $k$ = 3000. MLP architecture consist of two fully connected layers with 64 hidden units each and ReLU is applied after each hidden layer. Adam optimizer with a learning rate of $1\times10^{-3}$ is used, where MSE loss evaluated the difference between predicted and true values during training. We use batch size of 64, and the model was trained for 20,000 epochs with an early stopping patience: 5,000 epochs. 
    \item VAE with XGBoost. The VAE encoder consists of three fully connected layers (32, 64, and 16 neurons) with a latent dimension of 3, followed by a decoder with a mirrored structure. ReLU activation is applied after each hidden layer. The model is trained using MSE reconstruction loss and KL divergence, optimized with Adam ($1\times10^{-3}$)  learning rate) for 100 epochs with early stopping (patience: 50 epochs). Latent features extracted from the trained VAE are used as input for XGBoost, which undergoes hyperparameter tuning with Optuna over 100 trials, optimizing tree depth (3$\sim$12), learning rate (0.01$\sim$0.3), and regularization terms. The final model is trained using histogram-based optimization (tree\_method='hist') with GPU acceleration.
    \item U-Net. The architecture of U-Net is as follows. \textit{Encoder}: 4 convolutional blocks with 64, 128, 256, and 512 filters, each block consisting of two $3\times3$ convolutions, batch normalization, and ReLU, followed by max pooling. \textit{Bottleneck}: A 1024-filter convolutional block. \textit{Decoder}: Four up-convolutional blocks for upsampling, each followed by concatenation with corresponding encoder features and convolutional layers. The number of filters reduces progressively from 512 to 64. \textit{Output layer}: A $1\times1$ convolutional layer maps the 64-filter feature map to a single output channel. Training settings was used as follows: Adam optimizer with a learning rate of $1\times 10^{-4}$ and weight decay of $1\times 10^{-5}$, Cyclic learning rate strategy with $base\_lr$ = 10$^{-5}$ and $max\_lr$ = 10$^{-3}$, batch size: 16, number of epoch: 20K with early stopping triggered after 5K epochs.
    \item U-Net++. The architecture of U-Net++ is as follows. \textit{Encoder Path}: Four levels of convolutional layers with increasing channels (64, 128, 256, 512), followed by max-pooling layers for down-sampling. \textit{Decoder Path}: Decoding layers with up-sampling (bilinear interpolation) and skip connections from corresponding encoder levels, progressively reducing the spatial dimension and channel count. \textit{Final layer}: A $1\times1$ convolution reduces the output to a single channel representing predicted topography. Training settings were the same as the original U-Net.
    \item U-Net3+. The architecture of U-Net3+ is as follows. \textit{Encoder Path}: Four levels of convolutional layers, where the number of channels increases progressively (64, 128, 256, 512). Each level consists of a convolutional block with a $3\times3$ kernel, batch normalization, and ReLU activation, followed by max-pooling layers for down-sampling. \textit{Decoder Path with Full-Scale Skip Connections}: The decoder utilizes full-scale skip connections, where features from all previous encoder levels are concatenated with the current decoder level after up-sampling (using bilinear interpolation). This structure allows multi-scale features to be effectively combined at each decoding stage. The number of channels is progressively reduced (512 to 256, 256 to 128, and 128 to 64) using $1\times1$ convolutional layers with batch normalization and ReLU activation. \textit{Final Layer}: A $1\times1$ convolutional layer reduces the output to a single channel representing the predicted bed topography. Training settings were the same as the original U-Net.
    \item Attention U-Net. The architecture of Attention U-Net is as follows. \textit{Encoder Path}: Four levels of convolutional layers with increasing channels (64, 128, 256, 512), where each level consists of two $3\times3$ convolutional layers followed by ReLU activation. Down-sampling is performed using max-pooling layers, progressively reducing the spatial dimensions while increasing feature depth. \textit{Decoder Path}: Symmetrical to the encoder, the decoder performs up-sampling using transposed convolution layers. Each up-sampled feature map is concatenated with the corresponding encoder feature map through skip connections, enhanced by Attention Blocks. The attention mechanism selectively focuses on relevant spatial regions by weighting the encoder features based on their interaction with the decoder features. \textit{Attention Mechanism}: The attention blocks compute weights using gating signals ($g$) and skip-connected encoder features ($x$), applying a learned attention mask to the encoder features. \textit{Final Layer}: A $1\times1$ convolutional layer reduces the output to a single channel representing the predicted topography. The training settings are the same as the original U-Net. 
    \end{itemize}

\subsection{Results}
In this section, we present the evaluation results of the proposed framework, including a quantitative comparison across multiple performance metrics, an ablation study to assess the contributions of key components, a qualitative comparison for visual analysis of the predictions, and spatial generalization analysis. All experiments were conducted on a PC with Linux OS, Intel Xeon 2.4GHz processor, 256GB RAM, and GPU RTX A4500.   

\subsection{Quantitative Comparison}
The proposed framework, \textbf{DeepTopoNet}, demonstrates superior performance across all sub-regions in the Upernavik Isstr{\o}m region. Comprehensive evaluation metrics, including MAE, RMSE, R$^{2}$, SSIM, and PSNR, consistently show that DeepTopoNet achieves the lowest errors and the highest similarity indices across all sub-regions. This highlights its robustness in accurately predicting bed topography, even in regions with varying radar data density. Table~\ref{result_quantitative} provides a detailed quantitative comparison of all methods across the four sub-regions.

Compared to \textbf{traditional interpolation methods}, such as IDW~\cite{johnston2001using} and RBF~\cite{heryudono2010radial}, DeepTopoNet delivers significantly improved performance. For instance, while IDW achieves moderate results in some sub-regions (e.g., SSIM of 0.87 in Sub-region II), its RMSE values remain substantially higher (e.g., 101.05 m in Sub-region I). Similarly, RBF fails to capture finer spatial details, resulting in higher errors across all metrics. In contrast, DeepTopoNet achieves considerably lower RMSE and higher SSIM, demonstrating its ability to model spatial variability more effectively than interpolation methods.

Against \textbf{pixel-based machine learning methods}, including RF~\cite{sekulic2020random}, MLP-PE~\cite{li2021learnable} and VAE+XGBoost~\cite{yi2023evaluating}, DeepTopoNet exhibits a significant advantage. RF struggles to capture complex spatial relationships, as reflected in its relatively high RMSE values (e.g., 99.87 m in Sub-region I) and low R$^{2}$ scores across sub-regions. MLP-PE, despite incorporating positional encoding, performs poorly with high errors and negative R$^{2}$ in some sub-regions (e.g., Sub-region III), indicating limited generalization. In contrast, DeepTopoNet effectively leverages spatial features and multimodal data, achieving an RMSE as low as 19.38 m and an SSIM of 0.94 in Sub-region I. VAE+XGBoost performs better than MLP-PE and RF but still falls short of DeepTopoNet in all sub-regions. While VAE+XGBoost benefits from variational autoencoder-based feature extraction and gradient-boosted decision trees for regression, its RMSE remains significantly higher than DeepTopoNet across all sub-regions (e.g., 108.54 m in Sub-region I vs. 19.38 m for DeepTopoNet). 

When compared to \textbf{patch-based deep learning models}, including U-Net~\cite{ronneberger2015u}, U-Net++~\cite{zhou2018unet++}, U-Net3+~\cite{huang2020unet}, and Attention U-Net~\cite{oktay2018attention}, DeepTopoNet shows a clear improvement. Among these baselines, Attention U-Net performs the best, achieving competitive results with an RMSE of 33.33 m and SSIM of 0.90 in Sub-region I. However, DeepTopoNet consistently surpasses Attention U-Net, achieving a much lower RMSE of 19.38 m and a higher SSIM of 0.94 in the same sub-region. This indicates that DeepTopoNet’s combination of enriched input features, dynamic loss balancing, and robust architecture allows it to better capture spatial relationships and generalize effectively across diverse sub-regions. In summary, DeepTopoNet’s consistent outperformance across all categories of baselines demonstrates its superior capability for accurately reconstructing subglacial topography, providing a robust solution for regions with complex and sparse observational data.

Table~\ref{tri_result} presents the relative difference (\%) in TRI for DeepTopoNet and baseline methods, assessing their ability to capture terrain variability. Traditional interpolation methods, such as IDW and RBF, perform well in less rugged areas like Sub-region IV but struggle in complex terrains like Sub-region II, where RBF exhibits a high TRI deviation of 57.02\%. Similarly, pixel-based models, including Random Forest and MLP-PE, show significant TRI deviations, with MLP-PE exceeding 883.57\% in Sub-region II and 1223.73\% in Sub-region IV, indicating their inability to effectively model spatial dependencies and terrain complexity.

\begin{table}[ht!]
\caption{Performance comparison of DeepTopoNet and baseline methods with respect to the relative difference (\%) in Terrain Ruggedness Index (TRI).}
    \label{tri_result}
    \centering
    \resizebox{0.48\textwidth}{!}{
    \begin{tabular}{lcccc}
    \hline
    Method & Sub-region I & Sub-region II & Sub-region III & Sub-region IV \\
    \hline\hline
    IDW~\cite{johnston2001using} & 25.71 & 32.73 & 8.13 & 3.55 \\
    RBF~\cite{heryudono2010radial} & 32.20 & 57.02 & 15.77 & 1.98 \\
    RF~\cite{sekulic2020random} & 85.33 & 214.49 & 35.03 & 26.07 \\
    MLP-PE~\cite{li2021learnable} & 451.44 & 883.57 & 583.15 & 1223.73 \\
    VAE+XGBoost~\cite{yi2023evaluating}&23.05&97.50&24.22&36.12\\
    U-Net~\cite{ronneberger2015u} &19.26 & 97.59 & 230.51 & 2.50 \\
    U-Net++~\cite{zhou2018unet++} & 9.42 & 59.34 & 150.98 & 6.74 \\
    U-Net3+~\cite{huang2020unet} & 56.56 & 72.90 & 143.71 & 5.37 \\
    Att. U-Net~\cite{oktay2018attention} & 18.05 & \textbf{25.11} & 18.06 & 10.68 \\
    \textbf{DeepTopoNet} & \textbf{6.58} & 30.32 & \textbf{8.01} & \textbf{1.20} \\
    \hline    
    \end{tabular}
    }
\end{table}

Among patch-based deep learning models, U-Net variants demonstrate varying performance, with U-Net++ performing well in Sub-regions I and IV but struggling in Sub-region III (150.98\%), suggesting difficulty in adapting to different terrain structures. Attention U-Net achieves strong results in Sub-region II (25.11\%) but is inconsistent across other regions. DeepTopoNet consistently achieves the lowest TRI differences in Sub-regions I, III, and IV (6.58\%, 8.01\%, and 1.20\%, respectively), demonstrating its robustness in preserving terrain ruggedness. While its performance in Sub-region II (30.32\%) leaves room for improvement, it remains competitive with the best-performing baselines. These results confirm that DeepTopoNet effectively generalizes across diverse landscapes while maintaining superior topographic accuracy.

\subsection{Ablation Study}
To further assess the contribution of each component in DeepTopoNet, we conducted an ablation study by systematically removing components to evaluate their impact on the model's performance. The study aimed to quantify how the exclusion of these components affects predictive accuracy. First, to examine the importance of using patches with stride, gradient features, and trend surface features, we removed all three components. In this case, the model is trained using only the original 5 raw features without applying stride during patch extraction. Second, to isolate the effect of stride, we trained the model on patches without stride but retained the gradient and trend surface features. The results of the ablation experiments are summarized in Table~\ref{ablation_result}.

\textbf{w/o stride, gradient ($\nabla$), and trend surface ($\textbf{t}$) features}. Removing these components leads to a substantial decline in performance. For instance, in Sub-region IV, RMSE increases drastically to 457.06 m, and R$^{2}$ becomes negative (e.g., -57.74), indicating a failure to capture meaningful spatial patterns. In other sub-regions, metrics such as SSIM and PSNR are lower compared to DeepTopoNet, emphasizing the importance of these features for accurate predictions. 

\textbf{w/o Stride, w/ gradient ($\nabla$), and trend surface ($\textbf{t}$) features}. Retaining gradient and trend surface features while removing stride significantly improves performance compared to the first scenario. For example, in Sub-region I, RMSE reduces to 36.26 m, and R$^{2}$ increases to 0.96, demonstrating that gradient and trend surface features are vital for capturing spatial variability. However, the performance still lags behind DeepTopoNet in certain metrics, such as RMSE in Sub-region III (27.55 m vs. 32.69 m for DeepTopoNet).

\textbf{w/o BedMachine Loss ($\mathcal{L}_{m}$)}. Removing the BedMachine Loss ($\mathcal{L}_{m}$) significantly degrades the model’s performance, highlighting its critical role in training DeepTopoNet. Across all sub-regions, the absence of $\mathcal{L}_{m}$ leads to substantial increases in RMSE, particularly in Sub-region IV, where RMSE jumps to 106.00 m and R$^{2}$ drops to -2.16, indicating a severe loss of predictive accuracy. Additionally, SSIM and PSNR values decline across all sub-regions, reflecting a reduction in structural similarity and topographic detail retention. The poor performance suggests that $\mathcal{L}_{m}$ plays a crucial role in constraining the predictions, ensuring that the model maintains coherence with the global topographic trends derived from BedMachine data.

\textbf{DeepTopoNet}. The full DeepTopoNet framework achieves the best results across most metrics and sub-regions. For example, in Sub-region I, RMSE is the lowest at 19.38 m, and SSIM reaches 0.94, indicating excellent structural similarity between predictions and ground truth. This highlights the effectiveness of combining stride, $\nabla$, $\textbf{t}$, and $\mathcal{L}_{m}$ for robust and accurate bed topography predictions. The ablation study shows that each component contributes significantly to the performance of DeepTopoNet. While $\nabla$ and $\textbf{t}$ play a key role in improving accuracy, the inclusion of stride further enhances the model's ability to generalize and capture spatial context, leading to superior performance across all sub-regions. Additionally, the $\mathcal{L}_{m}$ proves to be a crucial component of the model. This confirms that $\mathcal{L}_{m}$ is essential for balancing high-resolution radar-derived observations with global-scale topographic trends, ensuring stability and accuracy in bed topography predictions. 

\begin{table}[]
\caption{Qualitative comparison of full grid predictions for Sub-region I. Each row shows the predicted topography, the corresponding BedMachine reference, and the difference, highlighting the performance of each method.}
    \label{qualitative}
    \centering
    \begin{tabular}{l}
    \hline
    Prediction  |  BedMachine~\cite{morlighem2017bedmachine,morlighem2020deep} | Difference\\
    \hline\hline
    IDW~\cite{johnston2001using}\\
    \begin{minipage}[c]{0.40\textwidth}\includegraphics[width=1\textwidth]{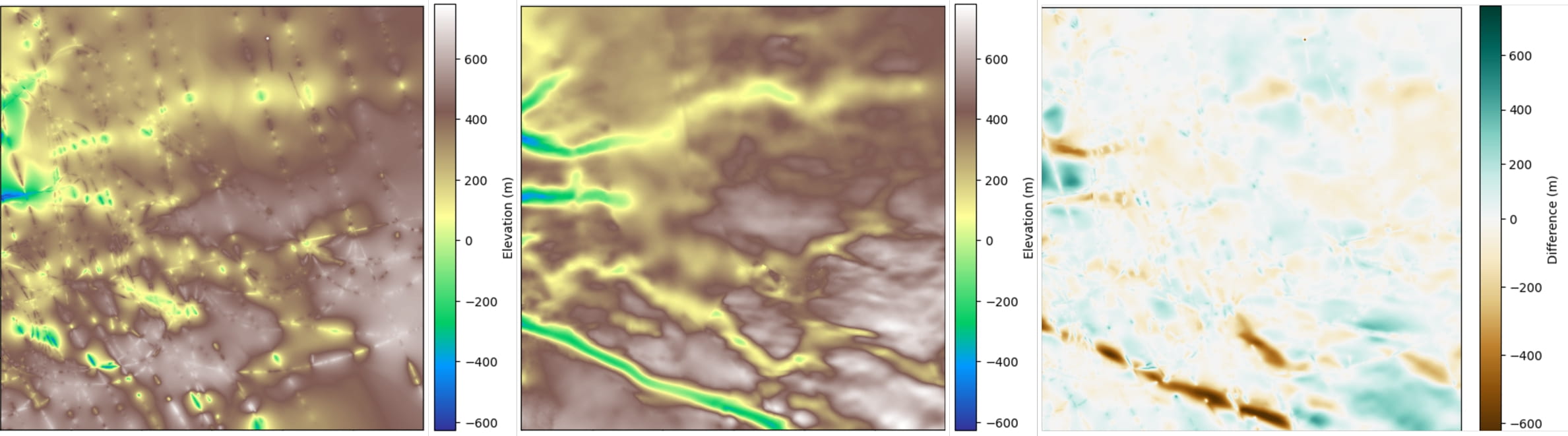}\end{minipage}\\
    RBF~\cite{heryudono2010radial}\\
    \begin{minipage}[c]{0.40\textwidth}\includegraphics[width=1\textwidth]{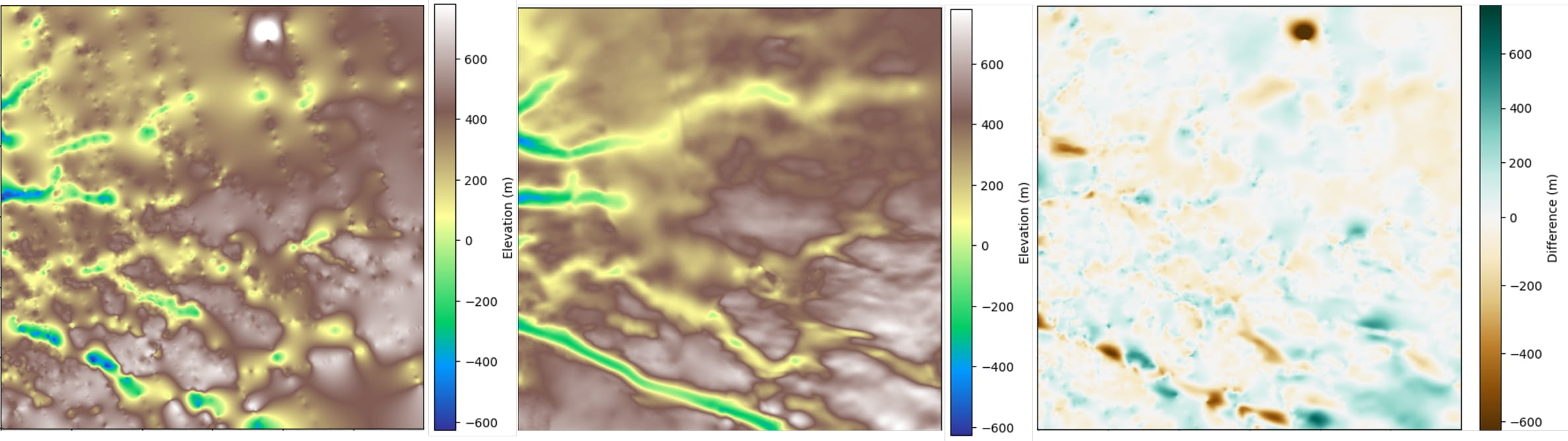}\end{minipage}\\ 
    RF~\cite{sekulic2020random}\\
    \begin{minipage}[c]{0.40\textwidth}\includegraphics[width=1\textwidth]{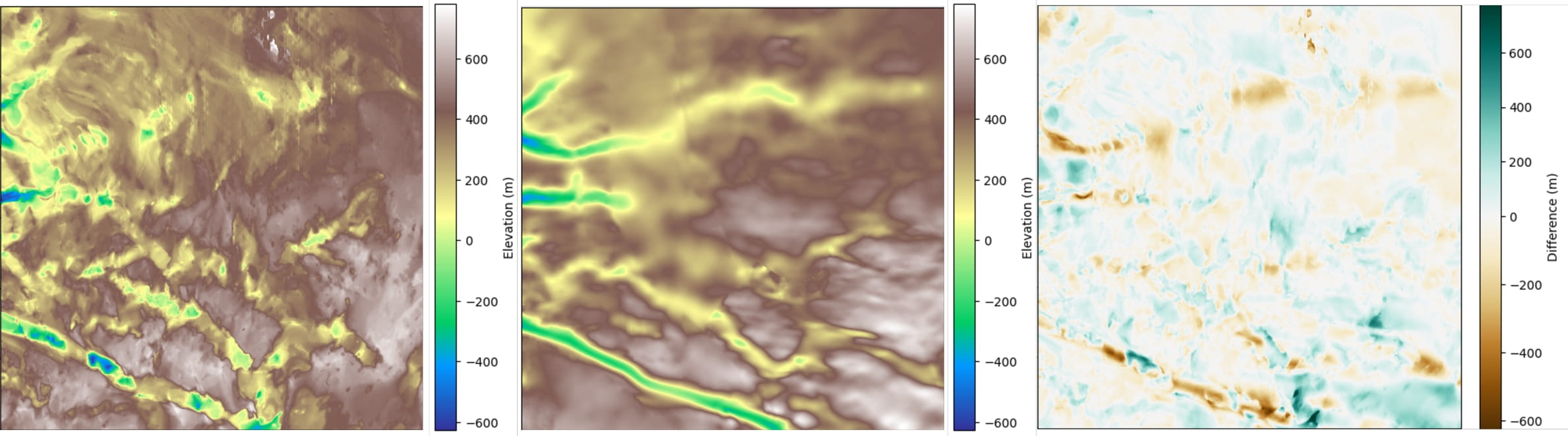}\end{minipage}\\    
    MLP-PE~\cite{li2021learnable}\\
    \begin{minipage}[c]{0.40\textwidth}\includegraphics[width=1\textwidth]{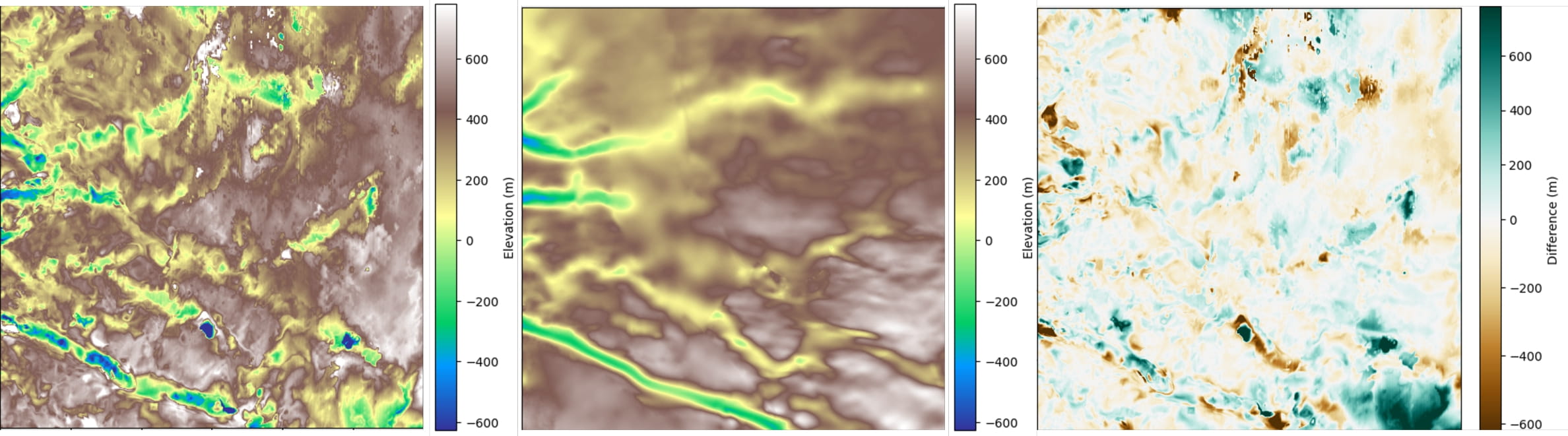}\end{minipage}\\
    VAE+XGBoost~\cite{yi2023evaluating}\\
    \begin{minipage}[c]{0.40\textwidth}\includegraphics[width=1\textwidth]{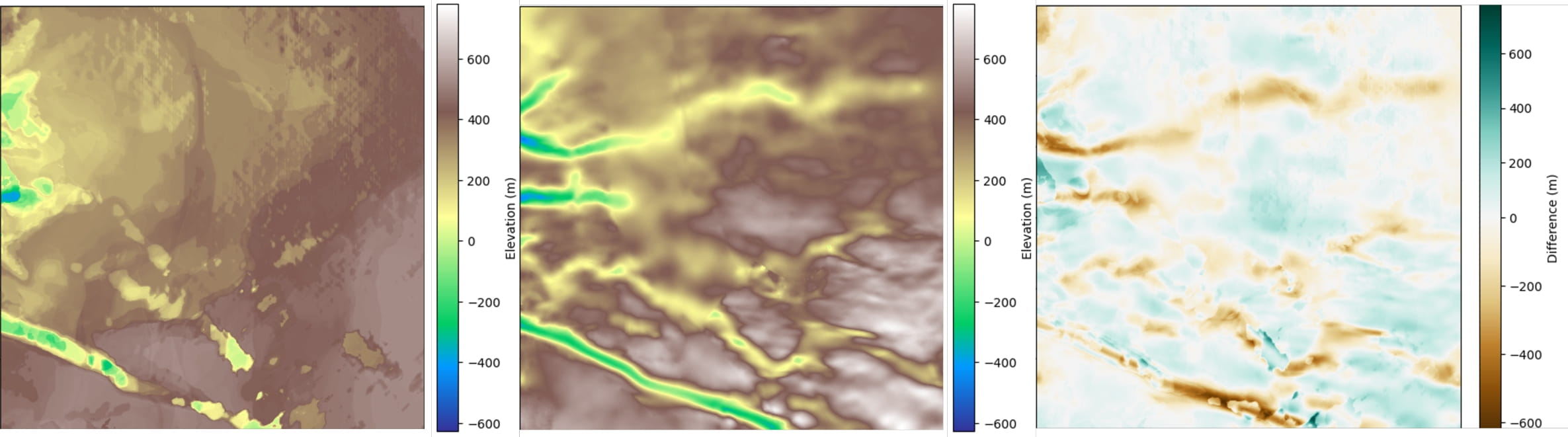}\end{minipage}\\
    U-Net~\cite{ronneberger2015u}\\
    \begin{minipage}[c]{0.40\textwidth}\includegraphics[width=1\textwidth]{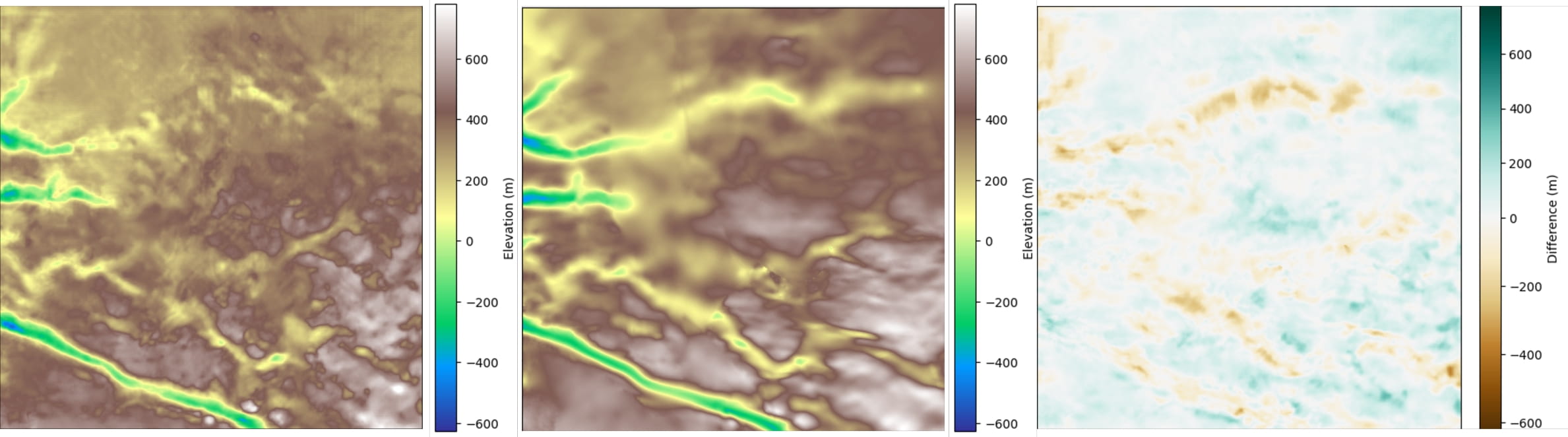}\end{minipage}\\
    U-Net++~\cite{zhou2018unet++}\\
    \begin{minipage}[c]{0.40\textwidth}\includegraphics[width=1\textwidth]{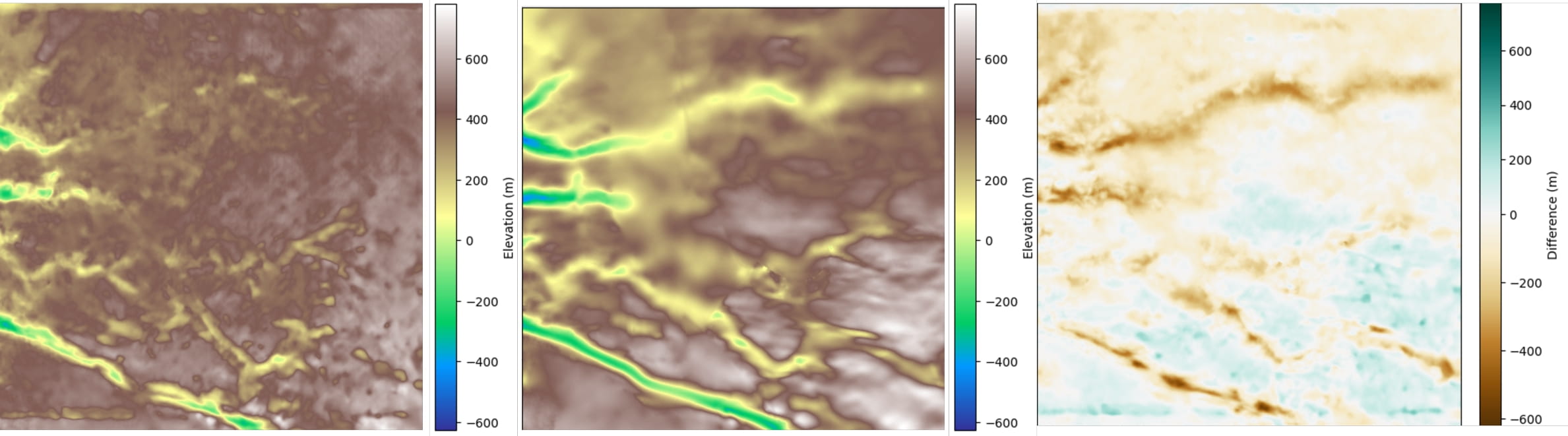}\end{minipage}\\
    U-Net3+~\cite{huang2020unet}\\
    \begin{minipage}[c]{0.40\textwidth}\includegraphics[width=1\textwidth]{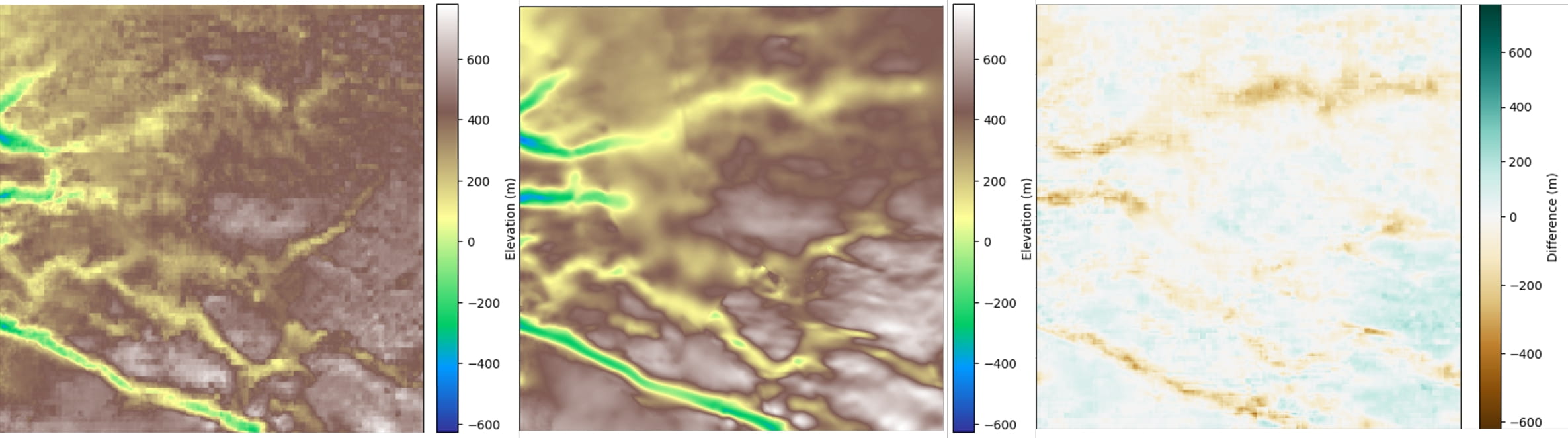}\end{minipage}\\  
    \\
    \hline
    \end{tabular}    
\end{table}

\begin{table}[]
\centering
\begin{tabular}{l}
\hline
Att. U-Net~\cite{oktay2018attention}\\
\begin{minipage}[c]{0.40\textwidth}\includegraphics[width=1\textwidth]{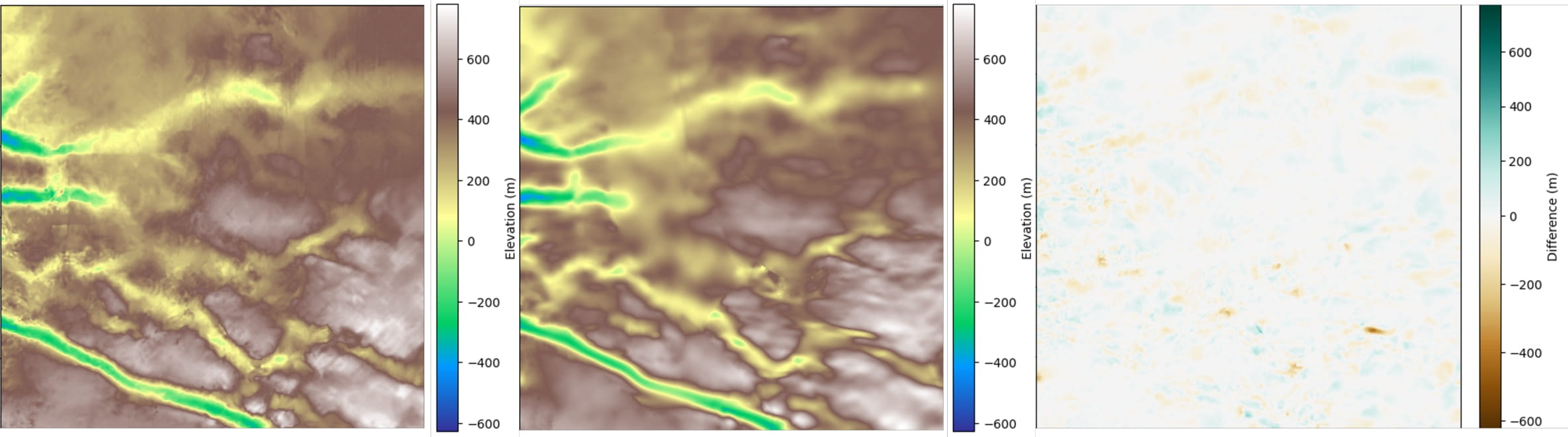}\end{minipage}\\    
\textbf{DeepTopoNet}\\
\begin{minipage}[c]{0.40\textwidth}\includegraphics[width=1\textwidth]{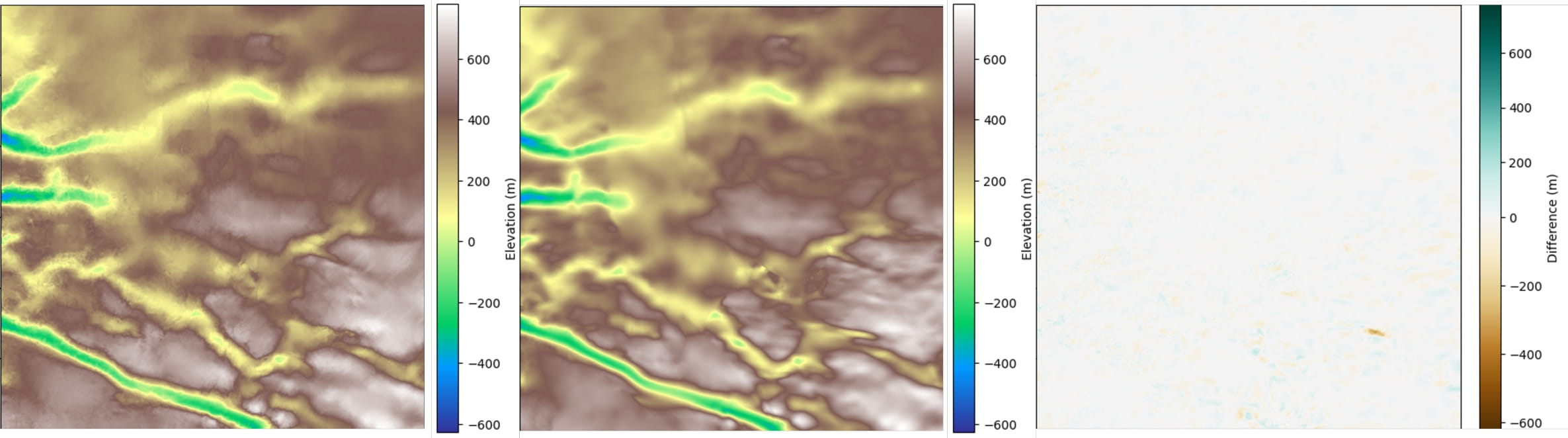}\end{minipage}\\
\\
\hline    
\end{tabular} 
\end{table}

\subsection{Qualitative Comparison}
The qualitative comparison in Table~\ref{qualitative} highlights the performance of various methods in predicting the full grid topography for Sub-region I. Each row presents the predicted topography, the reference BedMachine map, and the difference map, offering a visual assessment of spatial accuracy and consistency with ground truth.

\textbf{Interpolation Methods (IDW and RBF)}. Both IDW and RBF provide smooth predictions but fail to capture finer spatial details, as evident from the difference maps. IDW tends to oversmooth areas with high variability, leading to notable discrepancies in regions of steep topography. Similarly, RBF produces smoother outputs but struggles with localized variations, particularly in areas influenced by sharp transitions in the reference map.

\textbf{Pixel-Based Methods (RF, MLP-PE and VAE+XGB)}. Random forest (RF) shows improvement in capturing localized features compared to interpolation methods, yet the difference map reveals persistent inaccuracies in spatial patterns. On the other hand, MLP with positional encoding (MLP-PE) fails to generalize effectively, producing artifacts and noticeable errors in the difference map, reflecting its limited ability to model complex spatial relationships. VAE+XGBoost, which leverages a VAE to extract latent spatial representations before using XGBoost for prediction, exhibits moderate performance in reconstructing bed topography. While the VAE efficiently encodes high-dimensional spatial patterns, the combination with XGBoost still struggles in areas with sharp terrain transitions, leading to noticeable deviations in the difference map. 

\textbf{Patch-Based Methods (U-Net Variants)}. Among the U-Net family, standard U-Net provides better results compared to U-Net++ and U-Net3+, as evident from the more refined predictions and reduced discrepancies in the difference maps. Attention U-Net further improves spatial accuracy by focusing on critical regions, yielding smaller differences in areas of complex topography. However, all these methods exhibit slight smoothing in regions with abrupt elevation changes.

\textbf{DeepTopoNet.} The proposed DeepTopoNet demonstrates the best qualitative performance, with predictions closely matching the BedMachine reference. The difference map for DeepTopoNet shows minimal discrepancies, especially in areas with high variability and sharp transitions, highlighting its superior capability in capturing complex spatial patterns. This can be attributed to its robust architecture, dynamic loss balancing, and the integration of gradient and trend surface features, which enable precise predictions. Overall, DeepTopoNet outperforms all baseline methods in terms of spatial accuracy and alignment with the BedMachine reference. Its ability to generalize effectively across the entire grid and resolve fine spatial details underscores its robustness and reliability for subglacial topography prediction.

\subsection{Spatial Generalization Analysis}
To assess spatial generalization, we implement a grid-based validation strategy by horizontally slicing the input domain into 30 equal-width bands~\cite{beery2018recognition,meyer2019importance}. Odd-numbered slices (1, 3, 5, ...) are used for training and even-numbered slices (2, 4, 6, ...) for testing. This
design mimics realistic extrapolation to unobserved transects, reflecting operational deployment where radar coverage is spatially sparse.

\begin{table}[ht!]
\caption{Performance comparison of DeepTopoNet and deep learning baselines under grid-based spatial slicing paradigm for Sub-region I.}
    \label{spatialval}
    \centering
    \resizebox{0.48\textwidth}{!}{
    \begin{tabular}{lllllll}
    \hline
    Model&MAE$\downarrow$&RMSE$\downarrow$&R$^{2}\uparrow$&SSIM$\uparrow$&PSNR$\uparrow$&TRI$\downarrow$\\
    \hline\hline
    U-Net& 65.92& 90.52& 0.72& 0.79& 22.86& \textbf{15.24}\\
    U-Net++& 107.97& 142.74& 0.31& 0.69& 18.90& 17.78\\
    U-Net3+& 66.68& 88.98& 0.73& 0.75& 23.01& 21.89\\
    Att. U-Net& 23.69& 35.47& 0.96& 0.89& 30.99& 16.67\\
    \textbf{DeepTopoNet}& \textbf{15.90}& \textbf{25.39}& \textbf{0.98}& \textbf{0.91}& \textbf{33.90}& 17.94\\
    \hline

    \end{tabular}
    }
    
\end{table}
We evaluate DeepTopoNet and representative deep learning baselines (e.g., U-Net, U-Net++, U-Net3+, and Attention U-Net) under this spatial holdout scheme. As shown in Table~\ref{spatialval}, DeepTopoNet achieves the best performance across all major evaluation metrics in Sub-region I. It produces the lowest MAE (15.90 m) and RMSE (25.39 m), while maintaining the highest R$^{2}$ (0.98), SSIM (0.91), and
PSNR (33.90). Notably, even though U-Net and U-Net3+ perform moderately well, they fall short in both structural similarity and magnitude-based errors. While Attention U-Net achieves competitive performance, DeepTopoNet still outperforms it in all categories except TRI, where U-Net slightly edges out others in preserving localized ruggedness. 

These results confirm that DeepTopoNet not only excels under
standard training-validation splits but also maintains robust predictive performance when generalized to spatially unseen regions. This strengthens its applicability to real-world scenarios where
radar data is geographically sparse and extrapolation is critical.

\section{Conclusion}
This paper introduces \textbf{DeepTopoNet}, a novel deep learning framework for predicting subglacial topography in regions with sparse radar data. By integrating multimodal features (e.g., gradients, trend surfaces) and employing a dynamic loss balancing mechanism, DeepTopoNet achieves superior accuracy in reconstructing Greenland's bed topography across diverse sub-regions. Evaluations show it outperforms traditional interpolation, pixel-based machine learning, and state-of-the-art patch-based deep learning methods. Ablation studies confirm the importance of key components, such as stride, gradient features, and BedMachine loss, in enhancing model performance. DeepTopoNet excels in capturing fine spatial details and maintaining accuracy in varying data densities, demonstrating robustness and scalability. This framework not only improves subglacial topography modeling but also offers a scalable solution for other regions, bridging observational gaps and advancing insights into ice dynamics and sea-level rise. Furthermore, to support real-world deployment, we evaluate its spatial robustness via a grid-based slicing protocol that simulates extrapolation to unmeasured regions. In our future work, we aim to enhance the generalizability of DeepTopNet to varying subglacial topography by incorporating graph neural networks (GNNs) or knowledge-informed priors, which are better suited for capturing spatial dependencies and irregular topographic structures.

\section*{Acknowledgment}
This research has been funded by the NSF HDR Institute for Harnessing Data and Model Revolution in the Polar Regions (iHARP), Award \#2118285. 

\bibliographystyle{ACM-Reference-Format}
\bibliography{bedmap.bib}


\begin{thebibliography}{58}


\ifx \showCODEN    \undefined \def \showCODEN     #1{\unskip}     \fi
\ifx \showISBNx    \undefined \def \showISBNx     #1{\unskip}     \fi
\ifx \showISBNxiii \undefined \def \showISBNxiii  #1{\unskip}     \fi
\ifx \showISSN     \undefined \def \showISSN      #1{\unskip}     \fi
\ifx \showLCCN     \undefined \def \showLCCN      #1{\unskip}     \fi
\ifx \shownote     \undefined \def \shownote      #1{#1}          \fi
\ifx \showarticletitle \undefined \def \showarticletitle #1{#1}   \fi
\ifx \showURL      \undefined \def \showURL       {\relax}        \fi
\providecommand\bibfield[2]{#2}
\providecommand\bibinfo[2]{#2}
\providecommand\natexlab[1]{#1}
\providecommand\showeprint[2][]{arXiv:#2}

\bibitem[Agterberg(2023)]%
        {agterberg2023trend}
\bibfield{author}{\bibinfo{person}{Frits Agterberg}.} \bibinfo{year}{2023}\natexlab{}.
\newblock \showarticletitle{Trend surface analysis}.
\newblock In \bibinfo{booktitle}{\emph{Encyclopedia of Mathematical Geosciences}}. \bibinfo{publisher}{Springer}, \bibinfo{pages}{1567--1575}.
\newblock


\bibitem[{\AA}kesson et~al\mbox{.}(2021)]%
        {aakesson2021future}
\bibfield{author}{\bibinfo{person}{Henning {\AA}kesson}, \bibinfo{person}{Mathieu Morlighem}, \bibinfo{person}{Matt O’Regan}, {and} \bibinfo{person}{Martin Jakobsson}.} \bibinfo{year}{2021}\natexlab{}.
\newblock \showarticletitle{Future projections of Petermann Glacier under ocean warming depend strongly on friction law}.
\newblock \bibinfo{journal}{\emph{Journal of Geophysical Research: Earth Surface}} \bibinfo{volume}{126}, \bibinfo{number}{6} (\bibinfo{year}{2021}), \bibinfo{pages}{e2020JF005921}.
\newblock


\bibitem[Akiba et~al\mbox{.}(2019)]%
        {akiba2019optuna}
\bibfield{author}{\bibinfo{person}{Takuya Akiba}, \bibinfo{person}{Shotaro Sano}, \bibinfo{person}{Toshihiko Yanase}, \bibinfo{person}{Takeru Ohta}, {and} \bibinfo{person}{Masanori Koyama}.} \bibinfo{year}{2019}\natexlab{}.
\newblock \showarticletitle{Optuna: A next-generation hyperparameter optimization framework}. In \bibinfo{booktitle}{\emph{Proceedings of the 25th ACM SIGKDD international conference on knowledge discovery \& data mining}}. \bibinfo{pages}{2623--2631}.
\newblock


\bibitem[Appleby et~al\mbox{.}(2020)]%
        {appleby2020kriging}
\bibfield{author}{\bibinfo{person}{Gabriel Appleby}, \bibinfo{person}{Linfeng Liu}, {and} \bibinfo{person}{Li-Ping Liu}.} \bibinfo{year}{2020}\natexlab{}.
\newblock \showarticletitle{Kriging convolutional networks}. In \bibinfo{booktitle}{\emph{Proceedings of the AAAI Conference on Artificial Intelligence}}, Vol.~\bibinfo{volume}{34}. \bibinfo{pages}{3187--3194}.
\newblock


\bibitem[Bamber et~al\mbox{.}(2013)]%
        {bamber2013new}
\bibfield{author}{\bibinfo{person}{Jonathan~L Bamber}, \bibinfo{person}{JA Griggs}, \bibinfo{person}{RTW Hurkmans}, \bibinfo{person}{JA Dowdeswell}, \bibinfo{person}{SP Gogineni}, \bibinfo{person}{Ian Howat}, \bibinfo{person}{Jeremie Mouginot}, \bibinfo{person}{John Paden}, \bibinfo{person}{Steven Palmer}, \bibinfo{person}{Eric Rignot}, {et~al\mbox{.}}} \bibinfo{year}{2013}\natexlab{}.
\newblock \showarticletitle{A new bed elevation dataset for Greenland}.
\newblock \bibinfo{journal}{\emph{The Cryosphere}} \bibinfo{volume}{7}, \bibinfo{number}{2} (\bibinfo{year}{2013}), \bibinfo{pages}{499--510}.
\newblock


\bibitem[Bamber et~al\mbox{.}(2001)]%
        {bamber2001new}
\bibfield{author}{\bibinfo{person}{Jonathan~L Bamber}, \bibinfo{person}{Russell~L Layberry}, {and} \bibinfo{person}{SP Gogineni}.} \bibinfo{year}{2001}\natexlab{}.
\newblock \showarticletitle{A new ice thickness and bed data set for the Greenland ice sheet: 1. Measurement, data reduction, and errors}.
\newblock \bibinfo{journal}{\emph{Journal of Geophysical Research: Atmospheres}} \bibinfo{volume}{106}, \bibinfo{number}{D24} (\bibinfo{year}{2001}), \bibinfo{pages}{33773--33780}.
\newblock


\bibitem[Barnes et~al\mbox{.}(2021)]%
        {barnes2021transferability}
\bibfield{author}{\bibinfo{person}{Jowan~M Barnes}, \bibinfo{person}{Thiago Dias~dos Santos}, \bibinfo{person}{Daniel Goldberg}, \bibinfo{person}{G~Hilmar Gudmundsson}, \bibinfo{person}{Mathieu Morlighem}, {and} \bibinfo{person}{Jan De~Rydt}.} \bibinfo{year}{2021}\natexlab{}.
\newblock \showarticletitle{The transferability of adjoint inversion products between different ice flow models}.
\newblock \bibinfo{journal}{\emph{The Cryosphere}} \bibinfo{volume}{15}, \bibinfo{number}{4} (\bibinfo{year}{2021}), \bibinfo{pages}{1975--2000}.
\newblock


\bibitem[Beery et~al\mbox{.}(2018)]%
        {beery2018recognition}
\bibfield{author}{\bibinfo{person}{Sara Beery}, \bibinfo{person}{Grant Van~Horn}, {and} \bibinfo{person}{Pietro Perona}.} \bibinfo{year}{2018}\natexlab{}.
\newblock \showarticletitle{Recognition in terra incognita}. In \bibinfo{booktitle}{\emph{Proceedings of the European conference on computer vision (ECCV)}}. \bibinfo{pages}{456--473}.
\newblock


\bibitem[Briggs(1974)]%
        {briggs1974machine}
\bibfield{author}{\bibinfo{person}{Ian~C Briggs}.} \bibinfo{year}{1974}\natexlab{}.
\newblock \showarticletitle{Machine contouring using minimum curvature}.
\newblock \bibinfo{journal}{\emph{Geophysics}} \bibinfo{volume}{39}, \bibinfo{number}{1} (\bibinfo{year}{1974}), \bibinfo{pages}{39--48}.
\newblock


\bibitem[Cai et~al\mbox{.}(2025)]%
        {cai2025siamese}
\bibfield{author}{\bibinfo{person}{Yiheng Cai}, \bibinfo{person}{Yanliang He}, \bibinfo{person}{Shinan Lang}, \bibinfo{person}{Xiangbin Cui}, \bibinfo{person}{Xiaoqing Zhang}, {and} \bibinfo{person}{Zijun Yao}.} \bibinfo{year}{2025}\natexlab{}.
\newblock \showarticletitle{Siamese topographic generation model: A deep learning model for generating Antarctic subglacial topography with fine details}.
\newblock \bibinfo{journal}{\emph{Computers \& Geosciences}} (\bibinfo{year}{2025}), \bibinfo{pages}{105857}.
\newblock


\bibitem[Canny(1986)]%
        {canny1986computational}
\bibfield{author}{\bibinfo{person}{John Canny}.} \bibinfo{year}{1986}\natexlab{}.
\newblock \showarticletitle{A computational approach to edge detection}.
\newblock \bibinfo{journal}{\emph{IEEE Transactions on pattern analysis and machine intelligence}} \bibinfo{number}{6} (\bibinfo{year}{1986}), \bibinfo{pages}{679--698}.
\newblock


\bibitem[Cheng et~al\mbox{.}(2024)]%
        {cheng2024forward}
\bibfield{author}{\bibinfo{person}{Gong Cheng}, \bibinfo{person}{Mathieu Morlighem}, {and} \bibinfo{person}{Sade Francis}.} \bibinfo{year}{2024}\natexlab{}.
\newblock \showarticletitle{Forward and inverse modeling of ice sheet flow using physics-informed neural networks: Application to Helheim Glacier, Greenland}.
\newblock \bibinfo{journal}{\emph{Journal of Geophysical Research: Machine Learning and Computation}} \bibinfo{volume}{1}, \bibinfo{number}{3} (\bibinfo{year}{2024}), \bibinfo{pages}{e2024JH000169}.
\newblock


\bibitem[Feng et~al\mbox{.}(2024)]%
        {feng2024critical}
\bibfield{author}{\bibinfo{person}{Haoan Feng}, \bibinfo{person}{Yunting Song}, {and} \bibinfo{person}{Leila De~Floriani}.} \bibinfo{year}{2024}\natexlab{}.
\newblock \showarticletitle{Critical Features Tracking on Triangulated Irregular Networks by a Scale-Space Method}. In \bibinfo{booktitle}{\emph{Proceedings of the 32nd ACM International Conference on Advances in Geographic Information Systems}}. \bibinfo{pages}{54--66}.
\newblock


\bibitem[He et~al\mbox{.}(2016)]%
        {he2016deep}
\bibfield{author}{\bibinfo{person}{Kaiming He}, \bibinfo{person}{Xiangyu Zhang}, \bibinfo{person}{Shaoqing Ren}, {and} \bibinfo{person}{Jian Sun}.} \bibinfo{year}{2016}\natexlab{}.
\newblock \showarticletitle{Deep residual learning for image recognition}. In \bibinfo{booktitle}{\emph{Proceedings of the IEEE conference on computer vision and pattern recognition}}. \bibinfo{pages}{770--778}.
\newblock


\bibitem[Heryudono and Driscoll(2010)]%
        {heryudono2010radial}
\bibfield{author}{\bibinfo{person}{Alfa~RH Heryudono} {and} \bibinfo{person}{Tobin~A Driscoll}.} \bibinfo{year}{2010}\natexlab{}.
\newblock \showarticletitle{Radial basis function interpolation on irregular domain through conformal transplantation}.
\newblock \bibinfo{journal}{\emph{Journal of Scientific Computing}}  \bibinfo{volume}{44} (\bibinfo{year}{2010}), \bibinfo{pages}{286--300}.
\newblock


\bibitem[Hore and Ziou(2010)]%
        {hore2010image}
\bibfield{author}{\bibinfo{person}{Alain Hore} {and} \bibinfo{person}{Djemel Ziou}.} \bibinfo{year}{2010}\natexlab{}.
\newblock \showarticletitle{Image quality metrics: PSNR vs. SSIM}. In \bibinfo{booktitle}{\emph{2010 20th international conference on pattern recognition}}. IEEE, \bibinfo{pages}{2366--2369}.
\newblock


\bibitem[Howat et~al\mbox{.}(2014)]%
        {howat2014greenland}
\bibfield{author}{\bibinfo{person}{Ian~M Howat}, \bibinfo{person}{A Negrete}, {and} \bibinfo{person}{Benjamin~E Smith}.} \bibinfo{year}{2014}\natexlab{}.
\newblock \showarticletitle{The Greenland Ice Mapping Project (GIMP) land classification and surface elevation data sets}.
\newblock \bibinfo{journal}{\emph{The Cryosphere}} \bibinfo{volume}{8}, \bibinfo{number}{4} (\bibinfo{year}{2014}), \bibinfo{pages}{1509--1518}.
\newblock


\bibitem[Huang et~al\mbox{.}(2020)]%
        {huang2020unet}
\bibfield{author}{\bibinfo{person}{Huimin Huang}, \bibinfo{person}{Lanfen Lin}, \bibinfo{person}{Ruofeng Tong}, \bibinfo{person}{Hongjie Hu}, \bibinfo{person}{Qiaowei Zhang}, \bibinfo{person}{Yutaro Iwamoto}, \bibinfo{person}{Xianhua Han}, \bibinfo{person}{Yen-Wei Chen}, {and} \bibinfo{person}{Jian Wu}.} \bibinfo{year}{2020}\natexlab{}.
\newblock \showarticletitle{Unet 3+: A full-scale connected unet for medical image segmentation}. In \bibinfo{booktitle}{\emph{ICASSP 2020-2020 IEEE international conference on acoustics, speech and signal processing (ICASSP)}}. IEEE, \bibinfo{pages}{1055--1059}.
\newblock


\bibitem[Hutchinson(1995)]%
        {hutchinson1995interpolating}
\bibfield{author}{\bibinfo{person}{Michael~F Hutchinson}.} \bibinfo{year}{1995}\natexlab{}.
\newblock \showarticletitle{Interpolating mean rainfall using thin plate smoothing splines}.
\newblock \bibinfo{journal}{\emph{International journal of geographical information systems}} \bibinfo{volume}{9}, \bibinfo{number}{4} (\bibinfo{year}{1995}), \bibinfo{pages}{385--403}.
\newblock


\bibitem[Jacobs(2005)]%
        {jacobs2005image}
\bibfield{author}{\bibinfo{person}{David Jacobs}.} \bibinfo{year}{2005}\natexlab{}.
\newblock \showarticletitle{Image gradients}.
\newblock \bibinfo{journal}{\emph{Class Notes for CMSC}}  \bibinfo{volume}{426} (\bibinfo{year}{2005}), \bibinfo{pages}{1--3}.
\newblock


\bibitem[Johnston et~al\mbox{.}(2001)]%
        {johnston2001using}
\bibfield{author}{\bibinfo{person}{Kevin Johnston}, \bibinfo{person}{Jay~M Ver~Hoef}, \bibinfo{person}{Konstantin Krivoruchko}, {and} \bibinfo{person}{Neil Lucas}.} \bibinfo{year}{2001}\natexlab{}.
\newblock \bibinfo{booktitle}{\emph{Using ArcGIS geostatistical analyst}}. Vol.~\bibinfo{volume}{380}.
\newblock \bibinfo{publisher}{Esri Redlands}.
\newblock


\bibitem[Joughin et~al\mbox{.}(2010)]%
        {joughin2010greenland}
\bibfield{author}{\bibinfo{person}{Ian Joughin}, \bibinfo{person}{Ben~E Smith}, \bibinfo{person}{Ian~M Howat}, \bibinfo{person}{Ted Scambos}, {and} \bibinfo{person}{Twila Moon}.} \bibinfo{year}{2010}\natexlab{}.
\newblock \showarticletitle{Greenland flow variability from ice-sheet-wide velocity mapping}.
\newblock \bibinfo{journal}{\emph{Journal of Glaciology}} \bibinfo{volume}{56}, \bibinfo{number}{197} (\bibinfo{year}{2010}), \bibinfo{pages}{415--430}.
\newblock


\bibitem[Kendall et~al\mbox{.}(2018)]%
        {kendall2018multi}
\bibfield{author}{\bibinfo{person}{Alex Kendall}, \bibinfo{person}{Yarin Gal}, {and} \bibinfo{person}{Roberto Cipolla}.} \bibinfo{year}{2018}\natexlab{}.
\newblock \showarticletitle{Multi-task learning using uncertainty to weigh losses for scene geometry and semantics}. In \bibinfo{booktitle}{\emph{Proceedings of the IEEE conference on computer vision and pattern recognition}}. \bibinfo{pages}{7482--7491}.
\newblock


\bibitem[Kipf and Welling(2017)]%
        {kipf1609semi}
\bibfield{author}{\bibinfo{person}{TN Kipf} {and} \bibinfo{person}{M Welling}.} \bibinfo{year}{2017}\natexlab{}.
\newblock \showarticletitle{Semi-Supervised Classification with Graph Convolutional Networks.}. In \bibinfo{booktitle}{\emph{ICLR}}.
\newblock


\bibitem[Lam(1983)]%
        {lam1983spatial}
\bibfield{author}{\bibinfo{person}{Nina Siu-Ngan Lam}.} \bibinfo{year}{1983}\natexlab{}.
\newblock \showarticletitle{Spatial interpolation methods: a review}.
\newblock \bibinfo{journal}{\emph{The American Cartographer}} \bibinfo{volume}{10}, \bibinfo{number}{2} (\bibinfo{year}{1983}), \bibinfo{pages}{129--150}.
\newblock


\bibitem[LeCun et~al\mbox{.}(2015)]%
        {lecun2015deep}
\bibfield{author}{\bibinfo{person}{Yann LeCun}, \bibinfo{person}{Yoshua Bengio}, {and} \bibinfo{person}{Geoffrey Hinton}.} \bibinfo{year}{2015}\natexlab{}.
\newblock \showarticletitle{Deep learning}.
\newblock \bibinfo{journal}{\emph{nature}} \bibinfo{volume}{521}, \bibinfo{number}{7553} (\bibinfo{year}{2015}), \bibinfo{pages}{436--444}.
\newblock


\bibitem[Leong and Horgan(2020)]%
        {leong2020deepbedmap}
\bibfield{author}{\bibinfo{person}{Wei~Ji Leong} {and} \bibinfo{person}{Huw~Joseph Horgan}.} \bibinfo{year}{2020}\natexlab{}.
\newblock \showarticletitle{DeepBedMap: a deep neural network for resolving the bed topography of Antarctica}.
\newblock \bibinfo{journal}{\emph{The Cryosphere}} \bibinfo{volume}{14}, \bibinfo{number}{11} (\bibinfo{year}{2020}), \bibinfo{pages}{3687--3705}.
\newblock


\bibitem[Li and Heap(2008)]%
        {li2008review}
\bibfield{author}{\bibinfo{person}{Jin Li} {and} \bibinfo{person}{Andrew~D Heap}.} \bibinfo{year}{2008}\natexlab{}.
\newblock \showarticletitle{A review of spatial interpolation methods for environmental scientists}.
\newblock  (\bibinfo{year}{2008}).
\newblock


\bibitem[Li et~al\mbox{.}(2023)]%
        {li2023rainfall}
\bibfield{author}{\bibinfo{person}{Jia Li}, \bibinfo{person}{Yanyan Shen}, \bibinfo{person}{Lei Chen}, {and} \bibinfo{person}{Charles Wang~Wai Ng}.} \bibinfo{year}{2023}\natexlab{}.
\newblock \showarticletitle{Rainfall spatial interpolation with graph neural networks}. In \bibinfo{booktitle}{\emph{International conference on database systems for advanced applications}}. Springer, \bibinfo{pages}{175--191}.
\newblock


\bibitem[Li et~al\mbox{.}(2000)]%
        {li2000comparison}
\bibfield{author}{\bibinfo{person}{Xin Li}, \bibinfo{person}{Guodong Cheng}, {and} \bibinfo{person}{Ling Lu}.} \bibinfo{year}{2000}\natexlab{}.
\newblock \showarticletitle{Comparison of spatial interpolation methods}.
\newblock \bibinfo{journal}{\emph{Advances in Earth science}} \bibinfo{volume}{15}, \bibinfo{number}{3} (\bibinfo{year}{2000}), \bibinfo{pages}{260}.
\newblock


\bibitem[Li et~al\mbox{.}(2021)]%
        {li2021learnable}
\bibfield{author}{\bibinfo{person}{Yang Li}, \bibinfo{person}{Si Si}, \bibinfo{person}{Gang Li}, \bibinfo{person}{Cho-Jui Hsieh}, {and} \bibinfo{person}{Samy Bengio}.} \bibinfo{year}{2021}\natexlab{}.
\newblock \showarticletitle{Learnable fourier features for multi-dimensional spatial positional encoding}.
\newblock \bibinfo{journal}{\emph{Advances in Neural Information Processing Systems}}  \bibinfo{volume}{34} (\bibinfo{year}{2021}), \bibinfo{pages}{15816--15829}.
\newblock


\bibitem[MacAyeal(1989)]%
        {macayeal1989large}
\bibfield{author}{\bibinfo{person}{Douglas~R MacAyeal}.} \bibinfo{year}{1989}\natexlab{}.
\newblock \showarticletitle{Large-scale ice flow over a viscous basal sediment: Theory and application to ice stream B, Antarctica}.
\newblock \bibinfo{journal}{\emph{Journal of Geophysical Research: Solid Earth}} \bibinfo{volume}{94}, \bibinfo{number}{B4} (\bibinfo{year}{1989}), \bibinfo{pages}{4071--4087}.
\newblock


\bibitem[Merriam and Lippert(1966)]%
        {merriam1966geologic}
\bibfield{author}{\bibinfo{person}{Daniel~F Merriam} {and} \bibinfo{person}{RH Lippert}.} \bibinfo{year}{1966}\natexlab{}.
\newblock \showarticletitle{Geologic model studies using trend-surface analysis}.
\newblock \bibinfo{journal}{\emph{The Journal of Geology}} \bibinfo{volume}{74}, \bibinfo{number}{3} (\bibinfo{year}{1966}), \bibinfo{pages}{344--357}.
\newblock


\bibitem[Meyer et~al\mbox{.}(2019)]%
        {meyer2019importance}
\bibfield{author}{\bibinfo{person}{Hanna Meyer}, \bibinfo{person}{Christoph Reudenbach}, \bibinfo{person}{Stephan W{\"o}llauer}, {and} \bibinfo{person}{Thomas Nauss}.} \bibinfo{year}{2019}\natexlab{}.
\newblock \showarticletitle{Importance of spatial predictor variable selection in machine learning applications--Moving from data reproduction to spatial prediction}.
\newblock \bibinfo{journal}{\emph{Ecological Modelling}}  \bibinfo{volume}{411} (\bibinfo{year}{2019}), \bibinfo{pages}{108815}.
\newblock


\bibitem[Millan et~al\mbox{.}(2022)]%
        {millan2022ice}
\bibfield{author}{\bibinfo{person}{Romain Millan}, \bibinfo{person}{J{\'e}r{\'e}mie Mouginot}, \bibinfo{person}{Antoine Rabatel}, {and} \bibinfo{person}{Mathieu Morlighem}.} \bibinfo{year}{2022}\natexlab{}.
\newblock \showarticletitle{Ice velocity and thickness of the world’s glaciers}.
\newblock \bibinfo{journal}{\emph{Nature Geoscience}} \bibinfo{volume}{15}, \bibinfo{number}{2} (\bibinfo{year}{2022}), \bibinfo{pages}{124--129}.
\newblock


\bibitem[Morland(1987)]%
        {morland1987unconfined}
\bibfield{author}{\bibinfo{person}{LW Morland}.} \bibinfo{year}{1987}\natexlab{}.
\newblock \showarticletitle{Unconfined ice-shelf flow}.
\newblock \bibinfo{journal}{\emph{Dynamics of the West Antarctic ice sheet}}  \bibinfo{volume}{4} (\bibinfo{year}{1987}), \bibinfo{pages}{99--116}.
\newblock


\bibitem[Morlighem et~al\mbox{.}(2020)]%
        {morlighem2020deep}
\bibfield{author}{\bibinfo{person}{Mathieu Morlighem}, \bibinfo{person}{Eric Rignot}, \bibinfo{person}{Tobias Binder}, \bibinfo{person}{Donald Blankenship}, \bibinfo{person}{Reinhard Drews}, \bibinfo{person}{Graeme Eagles}, \bibinfo{person}{Olaf Eisen}, \bibinfo{person}{Fausto Ferraccioli}, \bibinfo{person}{Ren{\'e} Forsberg}, \bibinfo{person}{Peter Fretwell}, {et~al\mbox{.}}} \bibinfo{year}{2020}\natexlab{}.
\newblock \showarticletitle{Deep glacial troughs and stabilizing ridges unveiled beneath the margins of the Antarctic ice sheet}.
\newblock \bibinfo{journal}{\emph{Nature geoscience}} \bibinfo{volume}{13}, \bibinfo{number}{2} (\bibinfo{year}{2020}), \bibinfo{pages}{132--137}.
\newblock


\bibitem[Morlighem et~al\mbox{.}(2014)]%
        {morlighem2014deeply}
\bibfield{author}{\bibinfo{person}{Mathieu Morlighem}, \bibinfo{person}{Eric Rignot}, \bibinfo{person}{Jeremie Mouginot}, \bibinfo{person}{Helene Seroussi}, {and} \bibinfo{person}{Eric Larour}.} \bibinfo{year}{2014}\natexlab{}.
\newblock \showarticletitle{Deeply incised submarine glacial valleys beneath the Greenland ice sheet}.
\newblock \bibinfo{journal}{\emph{Nature Geoscience}} \bibinfo{volume}{7}, \bibinfo{number}{6} (\bibinfo{year}{2014}), \bibinfo{pages}{418--422}.
\newblock


\bibitem[Morlighem et~al\mbox{.}(2011)]%
        {morlighem2011mass}
\bibfield{author}{\bibinfo{person}{Mathieu Morlighem}, \bibinfo{person}{E Rignot}, \bibinfo{person}{H{\'e}lene Seroussi}, \bibinfo{person}{Eric Larour}, \bibinfo{person}{H Ben~Dhia}, {and} \bibinfo{person}{Denis Aubry}.} \bibinfo{year}{2011}\natexlab{}.
\newblock \showarticletitle{A mass conservation approach for mapping glacier ice thickness}.
\newblock \bibinfo{journal}{\emph{Geophysical Research Letters}} \bibinfo{volume}{38}, \bibinfo{number}{19} (\bibinfo{year}{2011}).
\newblock


\bibitem[Morlighem et~al\mbox{.}(2017)]%
        {morlighem2017bedmachine}
\bibfield{author}{\bibinfo{person}{Mathieu Morlighem}, \bibinfo{person}{Chris~N Williams}, \bibinfo{person}{Eric Rignot}, \bibinfo{person}{Lu An}, \bibinfo{person}{Jan~Erik Arndt}, \bibinfo{person}{Jonathan~L Bamber}, \bibinfo{person}{Ginny Catania}, \bibinfo{person}{Nolwenn Chauch{\'e}}, \bibinfo{person}{Julian~A Dowdeswell}, \bibinfo{person}{Boris Dorschel}, {et~al\mbox{.}}} \bibinfo{year}{2017}\natexlab{}.
\newblock \showarticletitle{BedMachine v3: Complete bed topography and ocean bathymetry mapping of Greenland from multibeam echo sounding combined with mass conservation}.
\newblock \bibinfo{journal}{\emph{Geophysical research letters}} \bibinfo{volume}{44}, \bibinfo{number}{21} (\bibinfo{year}{2017}), \bibinfo{pages}{11--051}.
\newblock


\bibitem[No{\"e}l et~al\mbox{.}(2018)]%
        {noel2018modelling}
\bibfield{author}{\bibinfo{person}{Brice No{\"e}l}, \bibinfo{person}{Willem~Jan Van De~Berg}, \bibinfo{person}{J~Melchior Van~Wessem}, \bibinfo{person}{Erik Van~Meijgaard}, \bibinfo{person}{DIrk Van~As}, \bibinfo{person}{Jan Lenaerts}, \bibinfo{person}{Stef Lhermitte}, \bibinfo{person}{Peter Kuipers~Munneke}, \bibinfo{person}{CJP Smeets}, \bibinfo{person}{Lambertus~H Van~Ulft}, {et~al\mbox{.}}} \bibinfo{year}{2018}\natexlab{}.
\newblock \showarticletitle{Modelling the climate and surface mass balance of polar ice sheets using RACMO2--Part 1: Greenland (1958--2016)}.
\newblock \bibinfo{journal}{\emph{The Cryosphere}} \bibinfo{volume}{12}, \bibinfo{number}{3} (\bibinfo{year}{2018}), \bibinfo{pages}{811--831}.
\newblock


\bibitem[Oktay et~al\mbox{.}(2018)]%
        {oktay2018attention}
\bibfield{author}{\bibinfo{person}{Ozan Oktay}, \bibinfo{person}{Jo Schlemper}, \bibinfo{person}{Loic~Le Folgoc}, \bibinfo{person}{Matthew Lee}, \bibinfo{person}{Mattias Heinrich}, \bibinfo{person}{Kazunari Misawa}, \bibinfo{person}{Kensaku Mori}, \bibinfo{person}{Steven McDonagh}, \bibinfo{person}{Nils~Y Hammerla}, \bibinfo{person}{Bernhard Kainz}, {et~al\mbox{.}}} \bibinfo{year}{2018}\natexlab{}.
\newblock \showarticletitle{Attention u-net: Learning where to look for the pancreas}.
\newblock \bibinfo{journal}{\emph{arXiv preprint arXiv:1804.03999}} (\bibinfo{year}{2018}).
\newblock


\bibitem[Raissi et~al\mbox{.}(2019)]%
        {raissi2019physics}
\bibfield{author}{\bibinfo{person}{Maziar Raissi}, \bibinfo{person}{Paris Perdikaris}, {and} \bibinfo{person}{George~E Karniadakis}.} \bibinfo{year}{2019}\natexlab{}.
\newblock \showarticletitle{Physics-informed neural networks: A deep learning framework for solving forward and inverse problems involving nonlinear partial differential equations}.
\newblock \bibinfo{journal}{\emph{Journal of Computational physics}}  \bibinfo{volume}{378} (\bibinfo{year}{2019}), \bibinfo{pages}{686--707}.
\newblock


\bibitem[Reily~Shawn et~al\mbox{.}(1999)]%
        {reily1999terrain}
\bibfield{author}{\bibinfo{person}{J Reily~Shawn}, \bibinfo{person}{D DeGloria~Stephen}, {and} \bibinfo{person}{A Elliot~Robert}.} \bibinfo{year}{1999}\natexlab{}.
\newblock \showarticletitle{Terrain Ruggedness Index That Quantifies Topographic Heterogeneity}.
\newblock \bibinfo{journal}{\emph{Intermountain Journal of Science}} \bibinfo{volume}{5}, \bibinfo{number}{1-4} (\bibinfo{year}{1999}), \bibinfo{pages}{23--27}.
\newblock


\bibitem[Riba et~al\mbox{.}(2020)]%
        {riba2020kornia}
\bibfield{author}{\bibinfo{person}{Edgar Riba}, \bibinfo{person}{Dmytro Mishkin}, \bibinfo{person}{Daniel Ponsa}, \bibinfo{person}{Ethan Rublee}, {and} \bibinfo{person}{Gary Bradski}.} \bibinfo{year}{2020}\natexlab{}.
\newblock \showarticletitle{Kornia: an open source differentiable computer vision library for pytorch}. In \bibinfo{booktitle}{\emph{Proceedings of the IEEE/CVF Winter Conference on Applications of Computer Vision}}. \bibinfo{pages}{3674--3683}.
\newblock


\bibitem[Ronneberger et~al\mbox{.}(2015)]%
        {ronneberger2015u}
\bibfield{author}{\bibinfo{person}{Olaf Ronneberger}, \bibinfo{person}{Philipp Fischer}, {and} \bibinfo{person}{Thomas Brox}.} \bibinfo{year}{2015}\natexlab{}.
\newblock \showarticletitle{U-net: Convolutional networks for biomedical image segmentation}. In \bibinfo{booktitle}{\emph{Medical image computing and computer-assisted intervention--MICCAI 2015: 18th international conference, Munich, Germany, October 5-9, 2015, proceedings, part III 18}}. Springer, \bibinfo{pages}{234--241}.
\newblock


\bibitem[Sekuli{\'c} et~al\mbox{.}(2020)]%
        {sekulic2020random}
\bibfield{author}{\bibinfo{person}{Aleksandar Sekuli{\'c}}, \bibinfo{person}{Milan Kilibarda}, \bibinfo{person}{Gerard~BM Heuvelink}, \bibinfo{person}{Mladen Nikoli{\'c}}, {and} \bibinfo{person}{Branislav Bajat}.} \bibinfo{year}{2020}\natexlab{}.
\newblock \showarticletitle{Random forest spatial interpolation}.
\newblock \bibinfo{journal}{\emph{Remote Sensing}} \bibinfo{volume}{12}, \bibinfo{number}{10} (\bibinfo{year}{2020}), \bibinfo{pages}{1687}.
\newblock


\bibitem[Sobel(1970)]%
        {sobel1970camera}
\bibfield{author}{\bibinfo{person}{Irwin~Edward Sobel}.} \bibinfo{year}{1970}\natexlab{}.
\newblock \bibinfo{booktitle}{\emph{Camera models and machine perception}}.
\newblock \bibinfo{publisher}{stanford university}.
\newblock


\bibitem[Soria et~al\mbox{.}(2023)]%
        {soria2023tiny}
\bibfield{author}{\bibinfo{person}{Xavier Soria}, \bibinfo{person}{Yachuan Li}, \bibinfo{person}{Mohammad Rouhani}, {and} \bibinfo{person}{Angel~D Sappa}.} \bibinfo{year}{2023}\natexlab{}.
\newblock \showarticletitle{Tiny and efficient model for the edge detection generalization}. In \bibinfo{booktitle}{\emph{Proceedings of the IEEE/CVF International Conference on Computer Vision}}. \bibinfo{pages}{1364--1373}.
\newblock


\bibitem[Team et~al\mbox{.}(2023)]%
        {ipcc2023}
\bibfield{editor}{\bibinfo{person}{Core~Writing Team}, \bibinfo{person}{H. Lee}, {and} \bibinfo{person}{J. Romero}} (Eds.). \bibinfo{year}{2023}\natexlab{}.
\newblock \bibinfo{booktitle}{\emph{Climate Change 2023: Synthesis Report}}.
\newblock \bibinfo{publisher}{Intergovernmental Panel on Climate Change (IPCC)}, \bibinfo{address}{Geneva, Switzerland}.
\newblock
\urldef\tempurl%
\url{https://www.ipcc.ch/report/ar6/syr/}
\showURL{%
\tempurl}


\bibitem[Team(2020)]%
        {imbie2020mass}
\bibfield{author}{\bibinfo{person}{The~IMBIE Team}.} \bibinfo{year}{2020}\natexlab{}.
\newblock \showarticletitle{Mass balance of the Greenland Ice Sheet from 1992 to 2018}.
\newblock \bibinfo{journal}{\emph{Nature}} \bibinfo{volume}{579}, \bibinfo{number}{7798} (\bibinfo{year}{2020}), \bibinfo{pages}{233--239}.
\newblock


\bibitem[Wackernagel and Wackernagel(2003a)]%
        {wackernagel2003ordinary}
\bibfield{author}{\bibinfo{person}{Hans Wackernagel} {and} \bibinfo{person}{Hans Wackernagel}.} \bibinfo{year}{2003}\natexlab{a}.
\newblock \showarticletitle{Ordinary kriging}.
\newblock \bibinfo{journal}{\emph{Multivariate geostatistics: an introduction with applications}} (\bibinfo{year}{2003}), \bibinfo{pages}{79--88}.
\newblock


\bibitem[Wackernagel and Wackernagel(2003b)]%
        {wackernagel2003universal}
\bibfield{author}{\bibinfo{person}{Hans Wackernagel} {and} \bibinfo{person}{Hans Wackernagel}.} \bibinfo{year}{2003}\natexlab{b}.
\newblock \showarticletitle{Universal kriging}.
\newblock \bibinfo{journal}{\emph{Multivariate Geostatistics: An Introduction with Applications}} (\bibinfo{year}{2003}), \bibinfo{pages}{300--307}.
\newblock


\bibitem[Wang et~al\mbox{.}(2023)]%
        {wang2023unraveling}
\bibfield{author}{\bibinfo{person}{Lijing Wang}, \bibinfo{person}{Luk Peeters}, \bibinfo{person}{Emma~J MacKie}, \bibinfo{person}{Zhen Yin}, {and} \bibinfo{person}{Jef Caers}.} \bibinfo{year}{2023}\natexlab{}.
\newblock \showarticletitle{Unraveling the uncertainty of geological interfaces through data-knowledge-driven trend surface analysis}.
\newblock \bibinfo{journal}{\emph{Computers \& Geosciences}}  \bibinfo{volume}{178} (\bibinfo{year}{2023}), \bibinfo{pages}{105419}.
\newblock


\bibitem[Wang et~al\mbox{.}(2021)]%
        {wang2021self}
\bibfield{author}{\bibinfo{person}{Ximei Wang}, \bibinfo{person}{Jinghan Gao}, \bibinfo{person}{Mingsheng Long}, {and} \bibinfo{person}{Jianmin Wang}.} \bibinfo{year}{2021}\natexlab{}.
\newblock \showarticletitle{Self-tuning for data-efficient deep learning}. In \bibinfo{booktitle}{\emph{International Conference on Machine Learning}}. PMLR, \bibinfo{pages}{10738--10748}.
\newblock


\bibitem[Wang et~al\mbox{.}(2004)]%
        {wang2004image}
\bibfield{author}{\bibinfo{person}{Zhou Wang}, \bibinfo{person}{Alan~C Bovik}, \bibinfo{person}{Hamid~R Sheikh}, {and} \bibinfo{person}{Eero~P Simoncelli}.} \bibinfo{year}{2004}\natexlab{}.
\newblock \showarticletitle{Image quality assessment: from error visibility to structural similarity}.
\newblock \bibinfo{journal}{\emph{IEEE transactions on image processing}} \bibinfo{volume}{13}, \bibinfo{number}{4} (\bibinfo{year}{2004}), \bibinfo{pages}{600--612}.
\newblock


\bibitem[Yi et~al\mbox{.}(2023)]%
        {yi2023evaluating}
\bibfield{author}{\bibinfo{person}{Katherine Yi}, \bibinfo{person}{Angelina Dewar}, \bibinfo{person}{Tartela Tabassum}, \bibinfo{person}{Jason Lu}, \bibinfo{person}{Ray Chen}, \bibinfo{person}{Homayra Alam}, \bibinfo{person}{Omar Faruque}, \bibinfo{person}{Sikan Li}, \bibinfo{person}{Mathieu Morlighem}, {and} \bibinfo{person}{Jianwu Wang}.} \bibinfo{year}{2023}\natexlab{}.
\newblock \showarticletitle{Evaluating Machine Learning and Statistical Models for Greenland Subglacial Bed Topography}. In \bibinfo{booktitle}{\emph{2023 International Conference on Machine Learning and Applications (ICMLA)}}. IEEE, \bibinfo{pages}{659--666}.
\newblock


\bibitem[Zhou et~al\mbox{.}(2018)]%
        {zhou2018unet++}
\bibfield{author}{\bibinfo{person}{Zongwei Zhou}, \bibinfo{person}{Md~Mahfuzur Rahman~Siddiquee}, \bibinfo{person}{Nima Tajbakhsh}, {and} \bibinfo{person}{Jianming Liang}.} \bibinfo{year}{2018}\natexlab{}.
\newblock \showarticletitle{Unet++: A nested u-net architecture for medical image segmentation}. In \bibinfo{booktitle}{\emph{Deep Learning in Medical Image Analysis and Multimodal Learning for Clinical Decision Support: 4th International Workshop, DLMIA 2018, and 8th International Workshop, ML-CDS 2018, Held in Conjunction with MICCAI 2018, Granada, Spain, September 20, 2018, Proceedings 4}}. Springer, \bibinfo{pages}{3--11}.
\newblock


\end{thebibliography}

\appendix
\section{Code Availability}
The source code for DeepTopoNet and all related experiments is available at: \texttt{https://github.com/bayuat/DeepTopoNet}

\section{Additional Results for Sub-region II, III, and IV}
To further evaluate the performance of the proposed framework and baseline methods, we present additional results for Sub-regions II, III, and IV (see Table~\ref{additional_result}). These regions provide diverse challenges due to variations in radar data density, terrain complexity, and topographic features. 

\begin{table}[b!]
\caption{Additional visual comparison of predicted topography across Sub-regions II, III, and IV.}
\label{additional_result}
    \centering
    \begin{tabular}{c}
    \hline
    Sub-region II | Sub-region III | Sub-region IV\\    
    \hline\hline
    IDW~\cite{johnston2001using}\\
    \begin{minipage}[c]{0.48\textwidth}\includegraphics[width=1\textwidth]{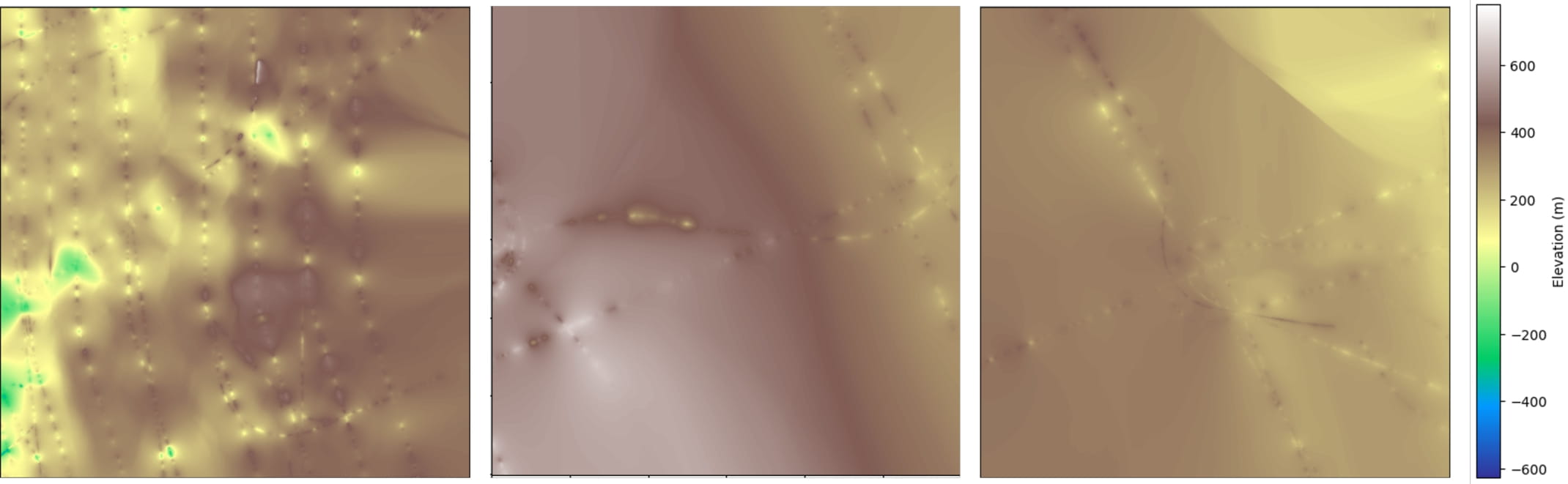}\end{minipage}\\       
    RBF~\cite{heryudono2010radial}\\
    \begin{minipage}[c]{0.48\textwidth}\includegraphics[width=1\textwidth]{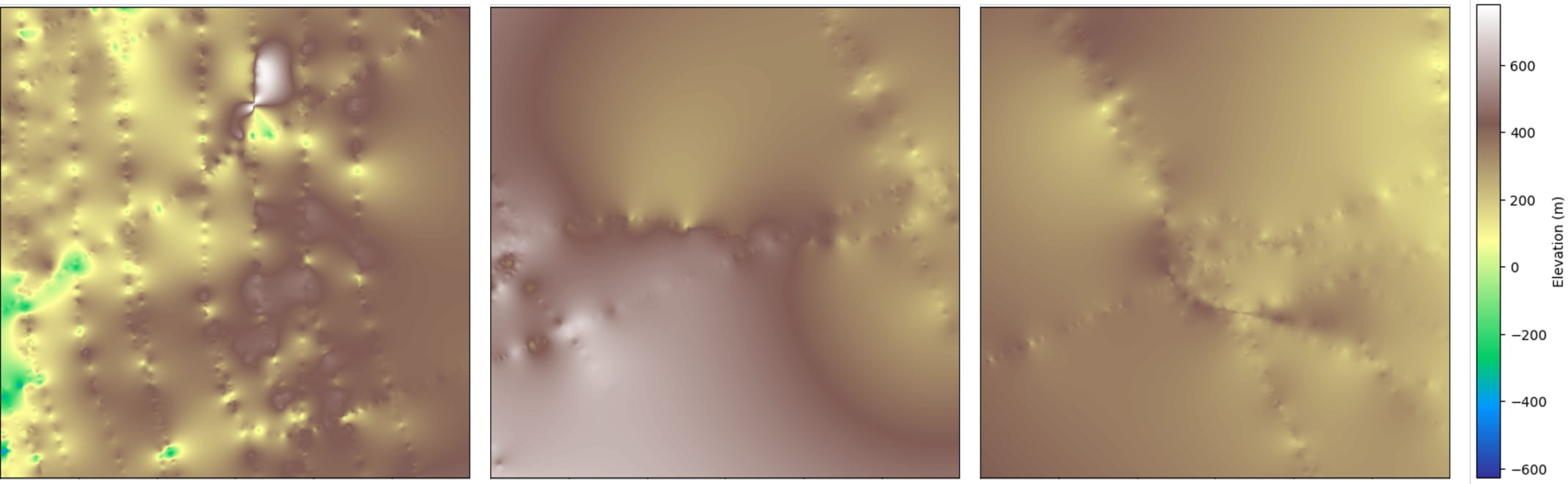}\end{minipage}\\
    RF~\cite{sekulic2020random}\\
    \begin{minipage}[c]{0.48\textwidth}\includegraphics[width=1\textwidth]{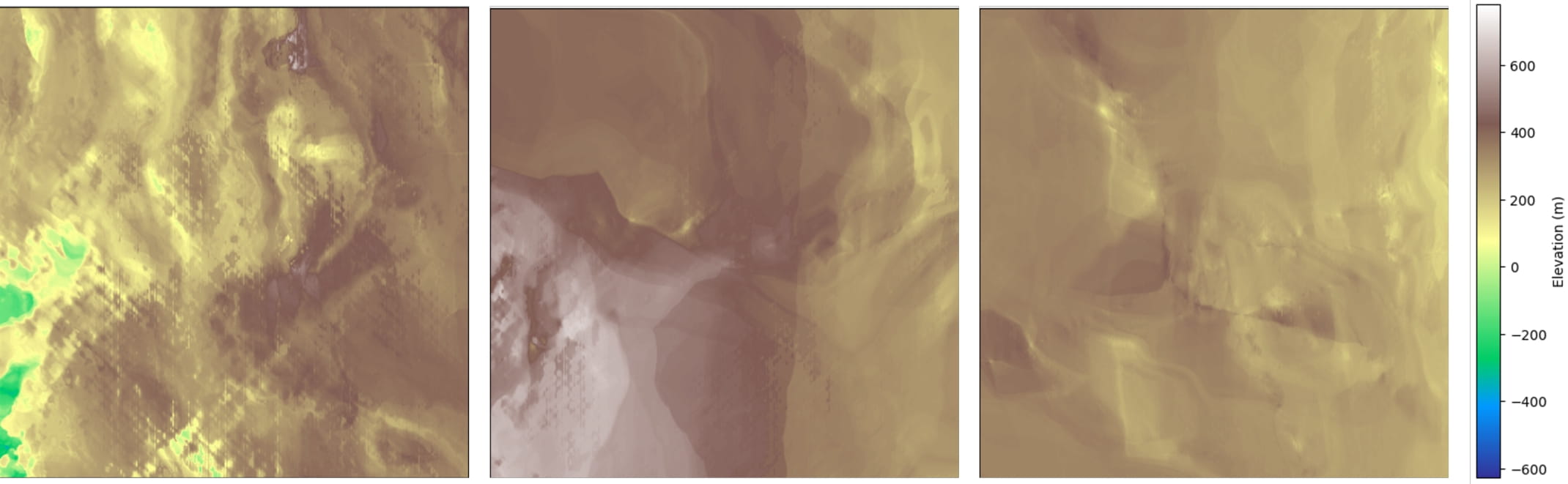}\end{minipage}\\
    MLP-PE~\cite{li2021learnable}\\ 
    \begin{minipage}[c]{0.48\textwidth}\includegraphics[width=1\textwidth]{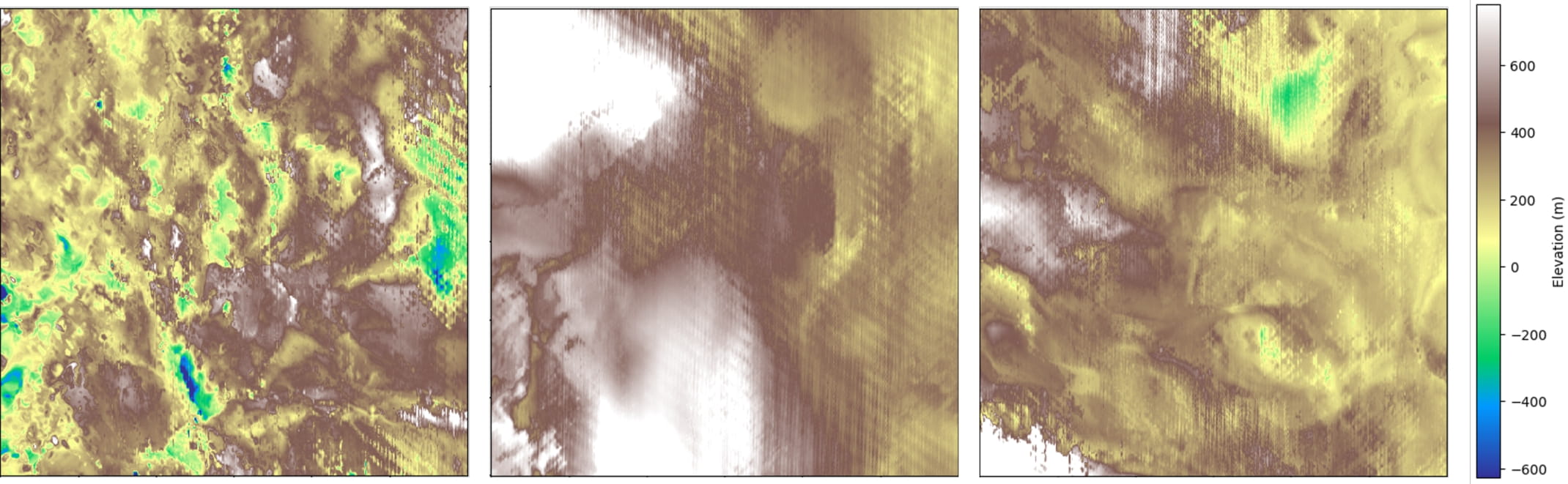}\end{minipage}\\
    VAE+XGBoost~\cite{yi2023evaluating}\\ 
    \begin{minipage}[c]{0.48\textwidth}\includegraphics[width=1\textwidth]{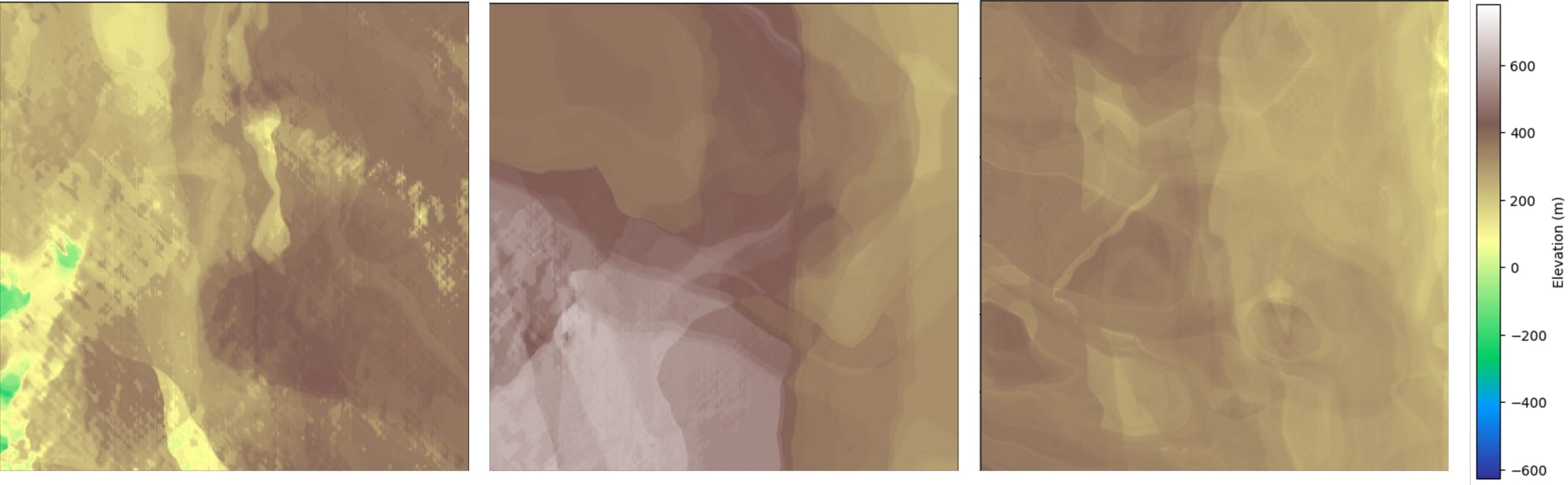}\end{minipage}\\
    U-Net~\cite{ronneberger2015u}\\
    \begin{minipage}[c]{0.48\textwidth}\includegraphics[width=1\textwidth]{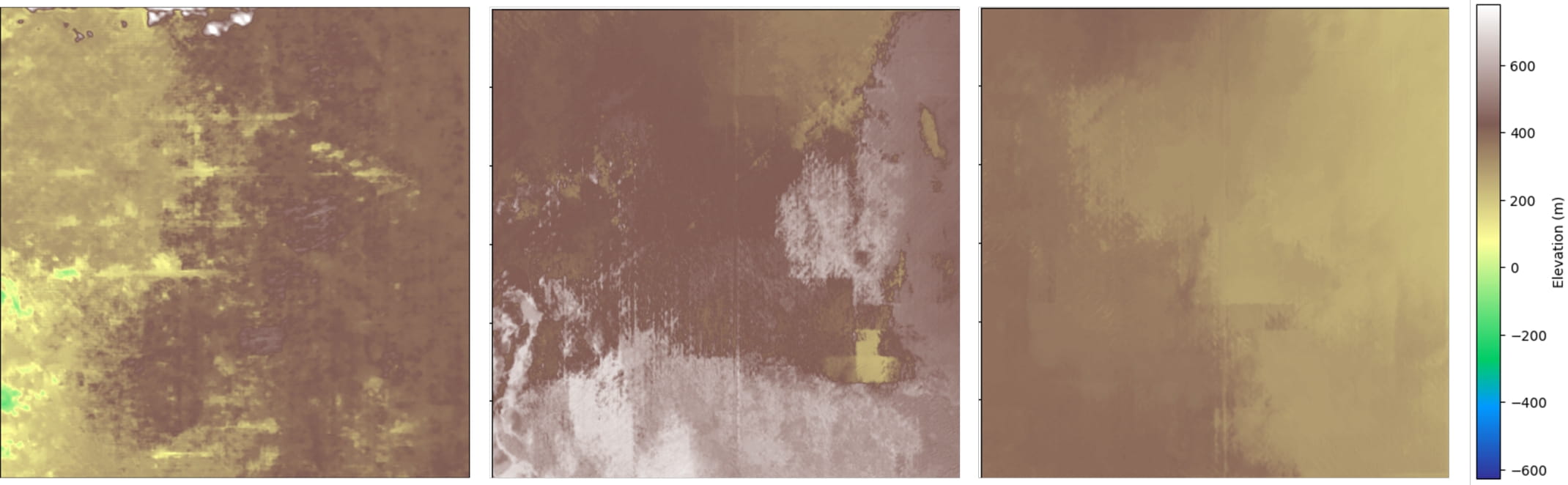}\end{minipage}\\
    \\
    \hline
    \end{tabular}    
\end{table}

\begin{table}[b!]
    \centering
    \begin{tabular}{c}
    \hline
    Sub-region II | Sub-region III | Sub-region IV\\    
    \hline\hline
    U-Net++~\cite{zhou2018unet++}\\
    \begin{minipage}[c]{0.48\textwidth}\includegraphics[width=1\textwidth]{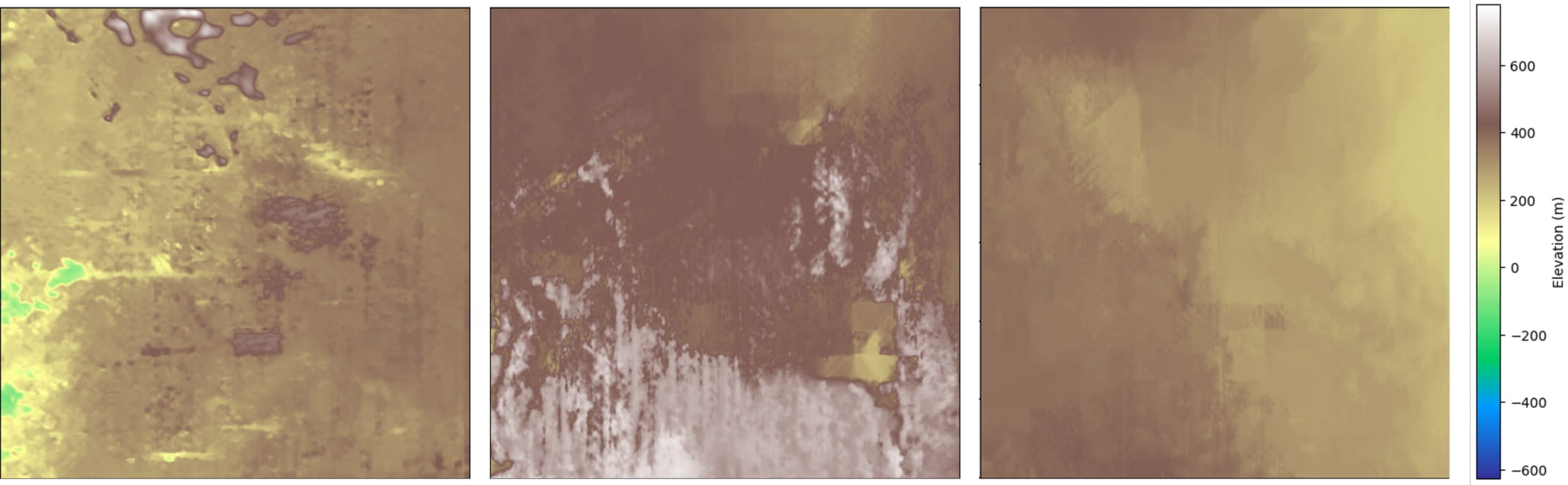}\end{minipage}\\
    U-Net3+~\cite{huang2020unet}\\
    \begin{minipage}[c]{0.48\textwidth}\includegraphics[width=1\textwidth]{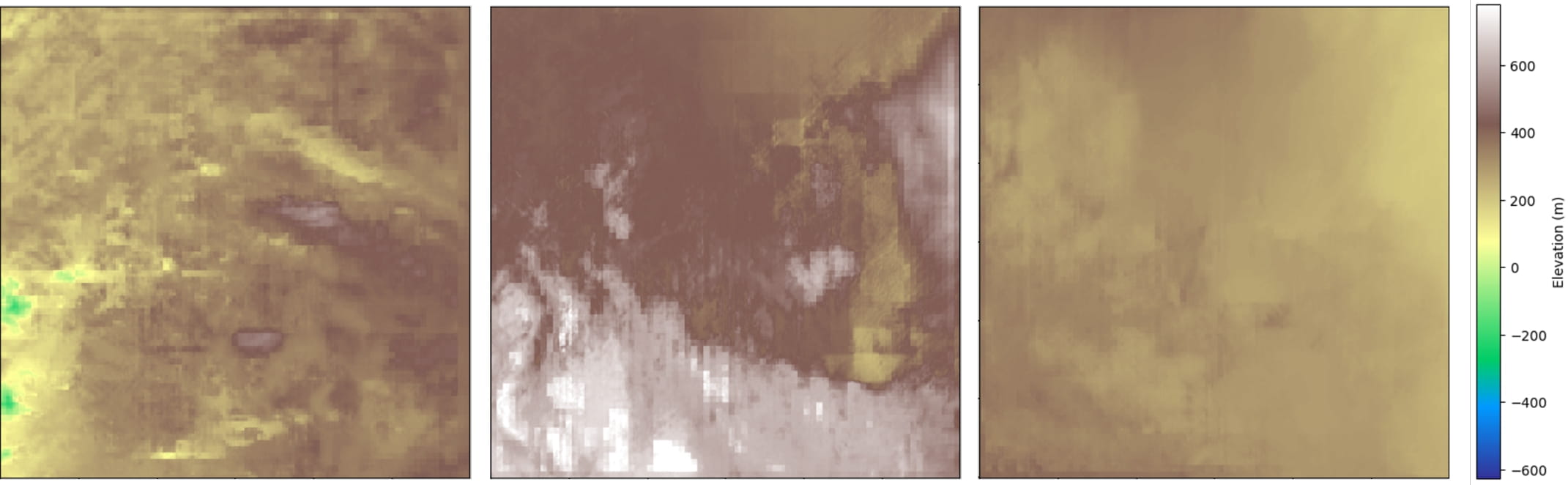}\end{minipage}\\
    Att. U-Net~\cite{oktay2018attention}\\
    \begin{minipage}[c]{0.48\textwidth}\includegraphics[width=1\textwidth]{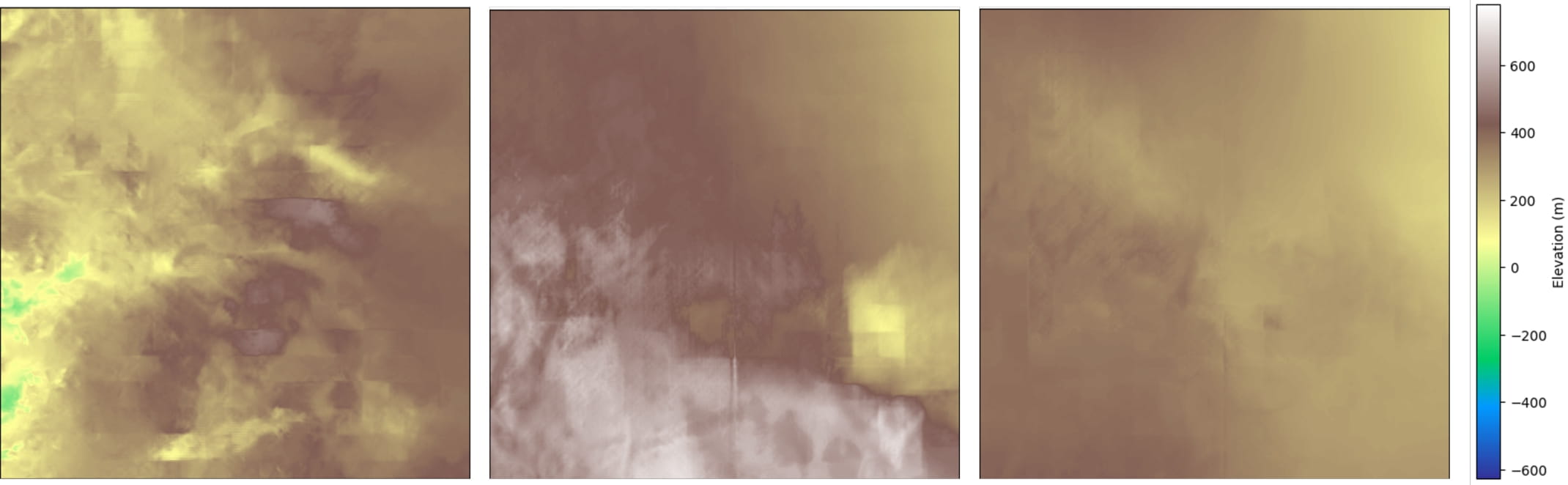}\end{minipage}\\
    \textbf{DeepTopoNet}\\
    \begin{minipage}[c]{0.48\textwidth}\includegraphics[width=1\textwidth]{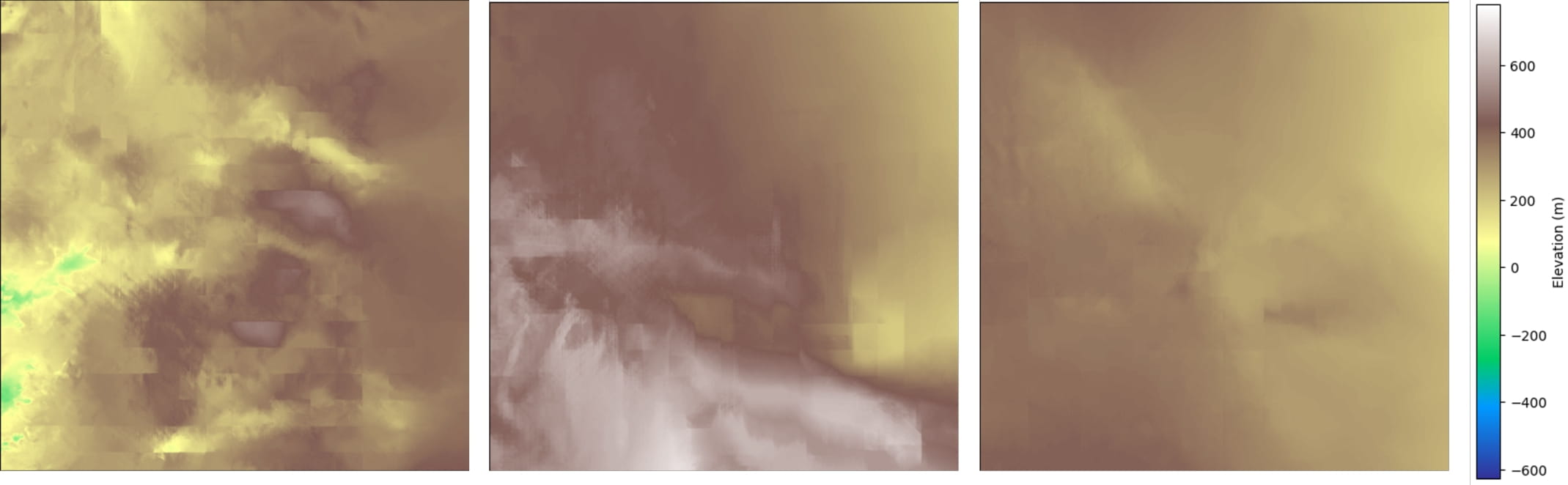}\end{minipage}\\    
    \\
    \hline
    \end{tabular}    
\end{table}
\end{document}